\documentclass[preprint,12pt]{elsarticle}

\usepackage{amssymb}
\usepackage{graphicx} 
\usepackage{amsmath, amssymb} 
\usepackage{float} 
\usepackage{hyperref}
\usepackage{setspace}
\usepackage{caption}
\usepackage{xcolor}
\journal{Mechanical Systems and Signal Processing}

\begin{document}

\begin{frontmatter}

\title{Equi-Euler GraphNet: An Equivariant, Temporal-Dynamics Informed Graph Neural Network for Dual Force and Trajectory Prediction in Multi-Body Systems}

\author[inst1]{Vinay Sharma}
\author[inst1]{Rémi Tanguy Oddon}
\affiliation[inst1]{organization={EPFL, Intelligent Maintenance and Operations Systems},
            city={Lausanne},
            country={Switzerland}}

\author[inst2]{Pietro Tesini}
\author[inst2]{Jens Ravesloot}
\author[inst2]{Cees Taal}
\author[inst1]{Olga Fink}
\affiliation[inst2]{organization={SKF, Research and Technology Development},
            city={Houten},
            country={The Netherlands}}

\begin{abstract}
Accurate real-time modeling of multi-body dynamical systems is essential for enabling  digital twin applications across industries. While  many data-driven approaches aim to learn system dynamics, jointly predicting internal loads and system trajectories remains a key challenge. This dual prediction  is particularly important for fault detection and predictive maintenance in continuously operating systems. Internal loads--such as contact forces--serve as early indicators of faults, reflecting changes due to wear, misalignment, or material degradation before these affect system motion. These forces also serve as inputs to physics-based degradation models (e.g., crack growth or wear propagation), enabling damage prediction and remaining useful life estimation, with the system state updated through predicted trajectories for continuous monitoring.

In this research, we propose Equi-Euler GraphNet, a physics-informed graph neural network (GNN) framework that simultaneously predicts internal forces and global trajectories in multi-body dynamical systems. 
In this mesh-free framework, nodes represent discrete components of the multi-body system and edges encode their interactions. Equi-Euler GraphNet introduces two novel inductive biases: (1) an \textbf{equivariant message-passing scheme} interpreting edge messages as interaction forces that remain consistent under Euclidean transformations; and (2) a \textbf{temporal-aware iterative node update mechanism based on Euler integration} that updates node states to capture the influence of distant interactions over time. Tailored  for cylindrical roller bearings, the framework introduces a structural bias that decouples the dynamics of the inner and outer rings from the constrained dynamics of the rolling elements. 

Trained on high-fidelity multiphysics simulations, Equi-Euler GraphNet extrapolates well beyond the training distribution, accurately predicting internal loads and system trajectories under extrapolated rotational  speeds, loads, and bearing configurations. It outperforms state-of-the-art GNN baselines focused solely on trajectory prediction, delivering stable rollouts over thousands of time steps with minimal error accumulation. Remarkably, it achieves up to a 200x speedup over conventional multiphysics solvers while maintaining comparable accuracy. These results position Equi-Euler GraphNet as an efficient reduced-order model for digital twins, design, and maintenance tasks.
\end{abstract}


\begin{highlights}
\item Novel physics-informed graph neural network framework.
\item Dual task of force and trajectory prediction in modeling of bearing dynamics.
\item Extrapolation to out-of-distribution operating conditions.
\end{highlights}


\end{frontmatter}



\section{Introduction}
\label{sec:introduction}
Multi-body dynamical systems are foundational to a wide range of engineered applications, including gearboxes, bearing assemblies, pumps, and compressors, where mechanical components interact to deliver the intended system   functionality. Accurately understanding and modeling these interactions is crucial not only for achieving  reliable performance  and meeting  design objectives, but also for enabling effective health monitoring, predicting  remaining useful life (RUL), and minimizing downtime due to unexpected failures \cite{cubillo2016review}. Traditional condition monitoring strategies typically rely on external observations, such as accelerometers mounted on housings, and infer system health by analyzing global vibration or acoustic signals  \cite{jardine2006review}. Although these condition monitoring technologies have demonstrated success in many industrial applications, they often only detect faults without estimating their severity or the fault location, nor do they capture the onset of damage early on. In multi-body systems, faults typically initiate at localized contact interfaces—such as gear meshing points \cite{li2012fatigue,pandya2013crack,li2017tribo} or rolling-element–raceway interfaces in bearings \cite{qiu2002damage,fajdiga2009fatigue,rycerz2017propagation}. By the time external vibration signals reveal such internal damage, the fault has often advanced to a more severe stage, leaving only a narrow window for preventive action \cite{de2016review}. Identifying and localizing faults at an earlier stage would enable more cost-effective refurbishment or partial reconditioning—particularly in large-scale systems like wind turbines—significantly reducing downtime and extending operational life.


In principle, internal loads at  contact interfaces provide  more direct and informative indicators  of damage initiation and progression \cite{fajdiga2009fatigue,morales2019}. Physically grounded damage models -- such as Paris’ law  for crack propagation \cite{pc1961rational} or wear models  for surface degradation \cite{meng1995wear} -- typically  take internal load histories  as key inputs, enabling accurate and long-range predictions of damage evolution. Accurately tracking these loads thus holds the potential for earlier detection of fault initiation and supports more precise remaining useful life prediction, facilitating timely and proactive maintenance scheduling. 

A prominent example  where accurate internal load estimation is particularly  critical is the rolling-element bearing, a fundamental component in nearly all rotating machinery across aerospace, automotive, and industrial applications. Bearings represent a particularly challenging modeling scenario  due to the complex, time-varying contact mechanics between rolling elements and raceways \cite{Lundberg1939,palmgren1959,Gupta1984}, which are further influenced by operating conditions such as rotational speed, external  loads, and lubrication conditions. However, direct measurement of internal  loads in systems such as bearings remains extremely challenging -- particularly in fast-rotating, confined, or lubricated environments. To overcome this limitation, researchers have turned to physics-based modeling approaches, most notably lumped-parameter models  \cite{cao2008comprehensive,cao2018mechanical,li2013tribo}  and finite element analysis (FEA) \cite{ku1998finite,wang2005finite}, to infer internal force distributions. These models integrate  available  information -- such as component 
 geometry, inertial  characteristics, and operating conditions -- into the governing equations of motion and contact mechanics, typically formulated as partial differential equations. Examples of physics-based condition monitoring include the use of dynamic bearing models to infer stiffness from measured vibration amplitudes and thereby estimate internal load distributions \cite{qiu2002damage}, as well as the integration of Paris’ law with FEA to track  crack growth in helicopter blades with vibration data as input \cite{kacprzynski2004predicting}.

Although physics-based approaches can provide  high-fidelity predictions of internal forces and resulting dynamics, their practical deployment in real-world settings remains limited. Lumped-parameter models often require extensive calibration to accurately reflect  system dynamics and are highly sensitive to input uncertainties; even minor  inaccuracies in estimated mass, stiffness, or damping can result in significant prediction errors \cite{liu2012data}. FEA-based methods, while  more comprehensive in modeling  contact mechanics and elastohydrodynamic phenomena,  are computationally intensive  and typically unsuitable  for continuous monitoring under varying  operating conditions \cite{peng2022digital}. Furthermore, as physical systems  deviate from the nominal assumptions -- whether through geometric tolerances, material inconsistencies, or fluctuating loads -- both lumped-parameter and FEA models often require re-parameterization or re-meshing, limiting  their adaptability for real-time or online applications.

Given  the limitations of traditional physics-based models, data-driven surrogate models have emerged as a promising  alternative \cite{choi2021data,zhu2021data,dimitrov2018wind}. By learning  directly on measured or simulated system data, these models can, in principle, approximate  the underlying dynamics without requiring explicit parameter calibration, and they tend to be more tolerant to noisy inputs -- a common challenge in industrial environments  \cite{an2015practical}. Typically,  such models generate trajectory predictions in an autoregressive manner -- commonly referred to as rollout -- where each predicted state is recursively used as input for the next  time step. However, purely data-driven approaches, including recurrent neural networks (RNNs) \cite{connor1994recurrent}, Long Short-Term Memory (LSTM) networks \cite{hochreiter1997long}, convolutional neural network (CNN) \cite{lecun1995convolutional}-based simulators, and other deep neural network architectures trained on historical system trajectories,  often struggle with error accumulation during extended rollouts \cite{karniadakis2021physics}, as they primarily learn statistical correlations within the training data rather than capturing true governing dynamics. This limitation restricts  their ability to generalize to unseen configurations or operating conditions. Physics-Informed Neural Networks (PINNs) \cite{raissi2019physics}  have been proposed to mitigate these shortcomings by incorporating  physical laws -- typically in the form of partial differential  equations -- directly into the training process as soft  constraints. This approach nudges the model toward physically consistent behavior. While  PINNs have shown promising results in domains such as fluid dynamics \cite{zhao2024comprehensive} and solid mechanics \cite{haghighat2021physics}, where the governing equations are well established and the  domains are relatively  continuous and  time-invariant,  their application to multi-body systems with time-varying contact interfaces  and boundaries presents substantial challenges \cite{new2023tunable,li2024physics}. In such settings, deriving a comprehensive set of coupled PDEs can be intractable, especially as the number and nature of contact interfaces evolve dynamically over time. Moreover, PINNs  are typically tailored to specific system geometries and boundary conditions, requiring  retraining for  variations such as different component layouts or added rolling elements. They are also sensitive to hyperparameter selection and problem-specific tuning \cite{krishnapriyan2021characterizing,new2023tunable},  which further constrains their scalability and reliability in  real-world industrial applications.

Graph Neural Networks (GNNs) \cite{battaglia2018relational} have emerged as a promising alternative for modeling complex physical systems by representing each system component as a node and their pairwise interactions -- such as contact forces, elastic couplings, or  damping effects -- as edges. This node–edge representation  introduces a spatial inductive bias that naturally aligns with the structure of multi-body systems  and offers  adaptability to changing  topologies, such as  when components are added, removed, or reconfigured \cite{battaglia2016interaction,mrowca2018flexible}. Early GNN-based approaches, such as the Graph Neural Simulator (GNS) \cite{sanchez2020learning}, have demonstrated  strong performance in learning trajectory evolutions for a range of physical systems, including granular flows \cite{choi2024graph} and molecular dynamics  \cite{atz2021geometric}. However, like  other data-driven methods, these approaches  are susceptible to error accumulation during  long-horizon rollouts, particularly when they lack embedded physical constraints \cite{thangamuthu2022unravelling}, which limits their ability to generalize to unseen regimes or operating conditions. To address these limitations, recent advances in equivariant GNNs, such as Equivariant Graph Neural Networks (EGNN) \cite{satorras2021n} and Graph Mechanics Networks (GMN) \cite{huang2022equivariant},  explicitly encode geometric symmetries -- such as rotation  and translation invariance -- into the model architecture. These models have shown improved performance in estimating physical quantities like velocity fields in N-body systems \cite{satorras2021n} and in tracking protein conformational dynamics  over time \cite{han2022equivariant}. Despite these advances, applying GNNs to  fault detection and internal load estimation presents a more complex challenge. This task requires the model to solve a dual objective: not only must  it accurately predict the evolution of node states (e.g., positions, velocities), but it must also accurately infer internal force distributions that drive damage initiation and progression. As we show in our experiments (Section \ref{sec:res_comp}), existing GNN methods -- including GNS \cite{sanchez2020learning}, EGNN \cite{satorras2021n}, GMN \cite{huang2022equivariant}) --  excel in trajectory prediction but fall short in delivering reliable estimates of  internal loads, a capability essential  for fault diagnosis  and  durability assessment in multi-body systems.


To address the limitations of existing GNN approaches in modeling multi-body systems, we propose the \textbf{Equi-Euler GraphNet} framework -- an equivariant  graph neural network approach specifically designed for predicting the internal loads and resulting system dynamics of  rolling-element bearings. This work builds on our previously proposed Dynami‑CAL GraphNet \cite{sharma2025dynami}, which used edge-local reference frames to conserve pair-wise linear and angular momentum and treated each message-passing step as a sub-time step integration. Equi-Euler GraphNet retains this sub-time-stepping formulation but replaces the expensive local frame construction with a lightweight, equivariant message-passing scheme, further tailored to the structural and kinematic characteristics of bearing systems.

Building on the strengths of prior  GNN architectures -- such as spatial inductive biases, symmetry preservation and sub time-stepping -- Equi-Euler GraphNet overcomes key limitations by explicitly targeting the accurate prediction of internal load distributions alongside dynamic trajectories. In this framework, each bearing component (rolling elements, inner ring, and outer ring) is represented as a node, while their physical interactions, including contact forces and ground reactions, are captured  through edge connections between these nodes.

A central  innovation of Equi-Euler GraphNet is its \textbf{equivariant message-passing mechanism}, which encodes physical interactions in a manner  consistent with Euclidean symmetries  (translations and rotations). During  each message-passing iteration, the model  performs two distinct yet  complementary computations through specialized  layers:

\begin{enumerate} \item A \textbf{dynamics prediction layer}, that estimates the internal forces acting on the inner and outer rings as a result of external loads,  and computes their corresponding accelerations. \item A \textbf{kinematics prediction layer} that computes the motion of the rolling elements, explicitly accounting for their constrained trajectories  between the rings. \end{enumerate}

These accelerations and displacements  are integrated over a small sub-time interval using a simple yet effective Euler integration scheme, enabling the update of nodal states (positions and velocities). Multiple message-passing  iterations are performed within each simulation time step, allowing the model to efficiently propagate force and motion information throughout the bearing assembly.

By repeatedly aggregating local interactions  across the graph, Equi-Euler GraphNet  is capable of accurately predicting both internal load distributions  and component-level trajectories over long simulation horizons, including rollouts spanning thousands of time steps -- an area where  existing GNN approaches often struggle due to error accumulation or lack of physical structure. Furthermore, the model exhibits strong generalization capabilities, reliably predicting bearing behavior  under a wide range of operating conditions and rotational speeds not encountered  during training. This robustness  reflects  the physically  grounded nature of the learned representations and underscores the model’s practical applicability in real-world industrial scenarios.

The key novelties and contributions of the proposed Equi-Euler Framework are summarized as follows:  

\begin{itemize}
    \item \textbf{Equivariant Message-Passing:}  
    We introduce a novel message-passing mechanism in which each edge message explicitly encodes a physical interaction force, ensuring consistency with  Euclidean transformations so that the predicted forces and trajectories remain consistent under rotation and translation, thereby improving generalization across operating regimes.  

    \item \textbf{Temporally aware Node Updates:}  
    Nodal states are iteratively updated using Euler integration at each message-passing step, embedding  a temporal inductive bias that accumulates the influence of  both local and distant  nodes over multiple sub-time steps. This structure  facilitates effective long-range interaction modeling and improves prediction accuracy over extended rollouts.      

    \item \textbf{Bearing-Specific Dual-Dynamics and Kinematics Layers Architecture:}  
We introduce a physically motivated architectural separation between dynamics and kinematics: a dynamics layer models the accelerations of the inner and outer rings in response to external load, while a dedicated  kinematics  layer captures the  constrained motion of the rolling elements between the rings. This decoupling  introduces a bearing-specific structural bias that improves  both trajectory prediction and internal load estimations.  
\end{itemize}

Consequently, Equi-Euler GraphNet reliably predicts both trajectories and internal force distributions  in bearing systems over extended simulation horizons. While our study focuses primarily on bearings, the generality of node-edge representation allows  the framework to be  readily applied  to  a broader class of  multi-body dynamical systems. Its dual prediction capability  -- capturing both loads and motions -- makes it particularly valuable for downstream tasks such as fault detection, condition monitoring, and damage evolution modeling within predictive maintenance applications. Moreover, the framework can efficiently generate realistic synthetic vibration data, helping to mitigate the pervasive issue of limited labeled datasets for supervised learning tasks  for bearing diagnostics \cite{wang2021}.

Beyond fault detection and health  modeling, Equi-Euler GraphNet offers significant  advantages for design optimization and system  evaluation. Trained on high-fidelity multiphysics  simulations that combine finite element analysis with elastohydrodynamic contact modeling, the framework  achieves a notable computational  speed-up: simulations  that take approximately two hours in the high-fidelity environment   can be completed in just 30 seconds using Equi-Euler GraphNet -- under comparable computational settings -- while maintaining trajectory accuracy. This efficiency  enables  rapid  exploration of diverse   bearing designs, loading conditions, and operating scenarios, making the framework  well-suited for parametric studies and sensitivity analyses. 

With its unique combination of predictive accuracy, computational efficiency, and generalization capability, Equi-Euler GraphNet provides a versatile and scalable  tool for accelerating the design, validation, and real-world deployment of complex multi-body dynamical systems across  industrial domains. 
The remainder of the paper is organized as follows: Section~\ref{sec:rel_work} reviews related work on GNN-based physical simulators and geometric inductive biases, focusing on state-of-the-art models relevant to our framework. Section~\ref{sec:problem} formalizes the modeling problem for cylindrical roller bearing systems, detailing the joint objectives of internal force estimation and long-horizon trajectory forecasting. Section \ref{sec:method} describes our proposed framework. First, the graph representation of a ground-mounted bearing is described (Section \ref{sec:graph_rep}) along with the node (Section \ref{sec:node_features}) and edge features (Section \ref{sec:edge_features}). Next, Section \ref{sec:model} introduces the Equi-Euler GraphNet single layer with its stacked dynamics predictor (Section \ref{sec:dyn_layer}) and kinematics predictor layers (Section \ref{sec:kin_layer}). The description of the framework is concluded with the description of the loss formulation (Section \ref{sec:loss}) and the testing pipeline (Section \ref{sec:test_pipeline}). In Section \ref{sec:case_study}, the case study used for evaluating the framework is described, detailing the multiphysics simulation setup for generating training and testing data. In Section~\ref{sec:comp_soa}, we describe the architectural differences among the GNN baselines used to benchmark the proposed method. Finally, Section \ref{sec:results} presents the results demonstrating Equi-Euler GraphNet's extrapolation performance and its comparison with baseline GNN models.

\section{Related Works}
\label{sec:rel_work}
This section highlights three state-of-the-art GNN architectures particularly relevant to our work: Graph Neural Simulator (GNS) \cite{sanchez2020learning}, Equivariant Graph Neural Network (EGNN) \cite{satorras2021n}, and Graph Mechanics Network (GMN) \cite{huang2022equivariant}.

GNS is one of the most prominent frameworks for simulating multi-body dynamical systems \cite{sanchez2020learning}. It models physical systems as graphs, with components represented as nodes and their pairwise interactions as edges. The architecture follows an encode–process–decode structure: the encoder embeds node states and relative positional features into latent representations; the processor performs multiple rounds of message passing to propagate interactions; and the decoder predicts particle-wise accelerations, which are integrated using a semi-implicit Euler scheme to update system states. While GNS benefits from strong spatial inductive biases -- enabling generalization across varying configurations -- it lacks explicit physical or geometric priors. As a result, it often accumulates errors during long-horizon rollouts and requires large amounts of training data to maintain  stable dynamic predictions. Moreover, GNS does not explicitly model internal forces as primary outputs. While forces can be inferred  from edge embeddings \cite{han2022learning}, their accuracy remains limited (see Section~\ref{sec:comp_soa} and Section \ref{sec:res_comp}).

To overcome  these limitations, recent approaches  have proposed geometric GNNs that incorporate symmetry-preserving inductive biases to ensure  physical consistency. These models explicitly distinguish between scalar and vector features and embed  geometric priors such as translational invariance and rotational equivariance. A notable  example is EGNN \cite{satorras2021n}, which adopts a scalarization–vectorization strategy: invariant scalar edge embeddings are first computed from pairwise node distances (scalarization), and these  are then used  to scale displacement vectors, generating  directional edge messages that remain equivariant under Euclidean transformations (vectorization). This architecture  improves generalization and stability while remaining  compact and computationally efficient.

Building on this foundation, GMN\cite{huang2022equivariant} extends EGNN by allowing each edge to carry multiple geometric vectors -- such as relative position, velocity, and other vectorial attributes -- structured as multi-channel representations. Within the scalarization–vectorization framework, scalar edge embeddings are computed using  normalized inner products of the stacked geometric vectors ensuring invariance to Euclidean transformations. These scalars are then decoded into weights that modulate  the original vectors, resulting in physically meaningful  and  directionally consistent edge messages. This design enables GMN to capture  more expressive and complex interactions while preserving geometric consistency.

While EGNN and GMN demonstrate improved trajectory prediction and robustness in physical simulations such as N-body and molecular dynamics, they do not explicitly  model internal forces as part of their core architecture. 


\section{Problem Formulation}
\label{sec:problem}
We consider a cylindrical roller bearing system \cite{Gupta1979} comprising rolling elements, an inner ring, and an outer ring supported by a stationary housing (ground). At each  discrete time step \(t\), the system state is defined  by the positions and velocities of  the center points of all  components:
\begin{equation}
\vec{\mathbf{x}}_i^{t}, \quad \vec{\mathbf{v}}_i^{t}, \quad i \in \{\text{Rolling Elements}, \text{Inner Ring}, \text{Outer Ring}\}.
\end{equation}

Given the system state at time \(t\), our objective is twofold:

\begin{enumerate}
\item \textbf{Instantaneous Force Prediction}: Predict the pairwise contact forces between interacting  components at time $t$ with high accuracy:
\begin{equation}
\vec{\mathbf{f}}_{ij}^{t}, \quad (i,j) \in 
\begin{aligned}
\{& \text{Rolling Element--Inner Ring}, \\
   & \text{Rolling Element--Outer Ring}, \\
   & \text{Inner Ring--Ground}, \\
   & \text{Outer Ring--Ground} \}.
\end{aligned}
\end{equation}
where, \(\vec{\mathbf{f}}_{ij}^{t}\) is the contact force.

\item \textbf{Trajectory Forecasting}: Predict the future state of each bearing component at a subsequent time step \( t + N \):
\begin{equation}
\vec{\mathbf{x}}_i^{t+N}, \quad \vec{\mathbf{v}}_i^{t+N}, \quad i \in \{\text{Rolling Elements}, \text{Inner Ring}, \text{Outer Ring}\}
\end{equation}
where, \(\vec{\mathbf{x}}_i^{t+N}, \quad \vec{\mathbf{v}}_i^{t+N}\) is the state of component \(\mathbf{i}\) at time \(t+N\)
\end{enumerate}

The task requires the model to learn the underlying system dynamics directly  from observed trajectory data, which consists of sampled  states and corresponding interaction forces at randomly selected  time steps -- without access to explicit physical  parameters, such as mass, damping, or stiffness matrices.  Training data is generated using  a high-fidelity multiphysics simulation environment, as illustrated  in Figure~\ref{fig:overview}(a). Alternatively, real-world measurements, such as  force signals obtained from  load cells embedded within the rolling elements, can also be used, similar to the measurement setup described  in \cite{zhao2025graph}. 

Physically, the dynamics of the bearing system are governed by a set of coupled partial differential equations that describe the evolution of the lubricant film and the elastic deformation of the contacting surfaces \cite{tsuha2020stiffness}. These elastohydrodynamic interactions result in highly  nonlinear and state-dependent behaviors, where effective mass, damping, and stiffness properties vary with operating conditions. Despite the underlying complexity, the global dynamics of the system can be approximated by a second-order differential equation of the form:
\begin{equation}
\textbf{M}\ddot{\vec{\mathbf{x}}} + \textbf{C}\dot{\vec{\mathbf{x}}} + \textbf{K}\vec{\mathbf{x}} = \vec{\mathbf{F}}_{\text{ext}}
\end{equation}
where \(\vec{\mathbf{x}}\) and \(\dot{\vec{\mathbf{x}}}\) denote the generalized position and velocity of the system, \(\textbf{M}\) is the mass matrix, and \(\textbf{C}\) and \(\textbf{K}\) represent the state-dependent effective damping and stiffness matrices that capture the internal forces generated through hydrodynamic contacts. The external forces are denoted by \(\vec{\mathbf{F}}_{\text{ext}}\). 

The trained model is tasked with implicitly learning the underlying  interactions and temporal dynamics -- predicting both internal forces (i.e. the combined effect of \(\textbf{K}\vec{\mathbf{x}}\) and \(\textbf{C}\dot{\vec{\mathbf{x}}}\)) and the resulting accelerations -- without direct access to the system parameters or the specific form of the \(\textbf{C}\), \(\textbf{K}\), and \(\textbf{M}\) matrices. Once trained, the model must perform recursive trajectory rollouts, predicting both future system states and internal forces over extended time horizons, as illustrated in Figure~\ref{fig:overview}(c). In addition, the model is expected to generalize effectively to unseen bearing configurations, rotational speeds, and external loads that fall well outside the range of training data. These capabilities  are critical  for  real-world deployment, where operating conditions are often variable and unpredictable, and where accurate long-term  forecasts are essential  for predictive maintenance, fault detection, and system reliability.


\section{Proposed Framework}
\label{sec:method}
\begin{figure}[h!]  
\centering \includegraphics[width=0.95\textwidth]{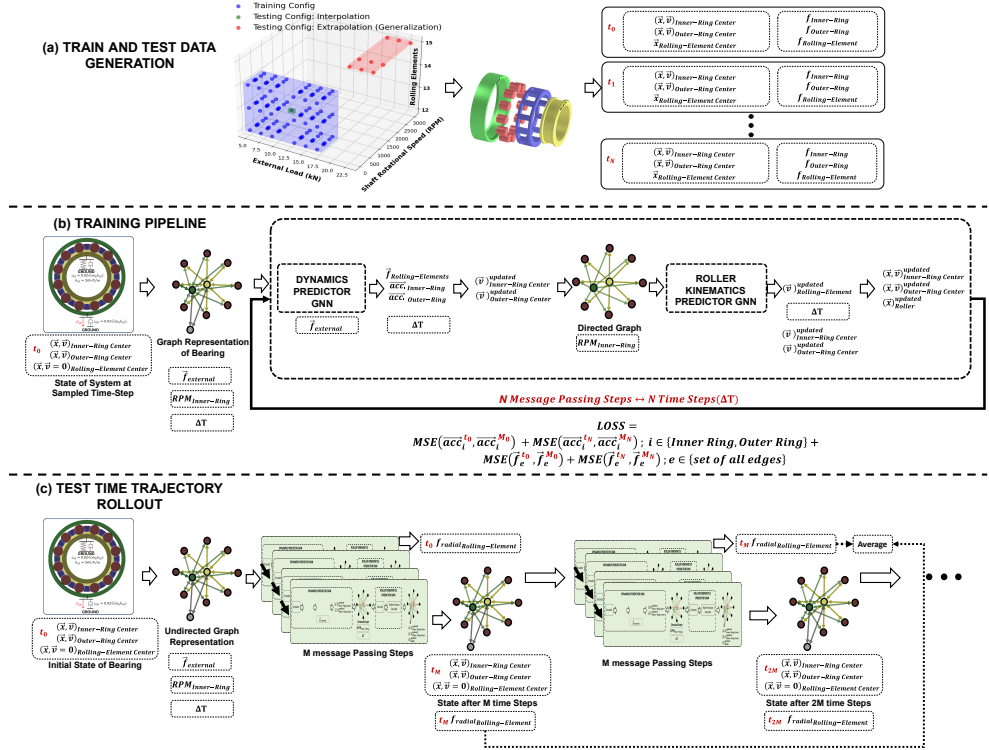}  
\caption{\textbf{Overview of Equi-Euler GraphNet Framework}:
\textbf{(a)} Train and test data generation from a multiphysics simulator, covering both interpolation and extrapolation cases.
\textbf{(b)} Stacked GNN pipeline with \emph{equivariant} message passing, comprising a \emph{dynamics prediction layer} (for ring accelerations and forces) and a \emph{kinematics prediction layer} (for roller motion), integrated over multiple sub-time steps using Euler integration. A loss function compares predicted accelerations and contact forces at initial and final sub-time steps.
\textbf{(c)} Test-time trajectory rollout with iterative state updates over extended horizons, enabling accurate long-term predictions under unseen operating conditions.}
\label{fig:overview}  
\end{figure}
In this section, we present  the proposed \textit{Equi-Euler GraphNet} framework for modeling the dynamics of cylindrical roller bearings. An overview of the model architecture is illustrated  in Figure~\ref{fig:overview}(b).

The framework adopts  a graph-based representation (Section \ref{sec:graph_rep}), where each bearing component -- including rolling elements, inner ring, outer ring, and ground -- is modeled as a distinct node type. Each node represents  a lumped abstraction    of its corresponding component, characterized   by the position and velocity of its center point, along with a scalar identity feature. This structure enables mesh-free, computationally efficient   modeling of complex bearing interactions (Section \ref{sec:node_features}). Pairwise physical interactions are represented  as edges that connect the nodes, with each edge carrying both vector and scalar features. Vector features encode  relative position and velocity, essential for capturing  physical dynamics, while scalar features define  the interaction type -- distinguishing, for example,  contact forces at rolling element-ring  interfaces from  reaction forces at component–ground boundaries (Section \ref{sec:edge_features}).

The core of \textit{Equi-Euler GraphNet} consists of two sequential predictor layers: the \textit{Dynamics Predictor Layer} (Section \ref{sec:dyn_layer}) and the \textit{Kinematics Predictor Layer} (Section \ref{sec:kin_layer}). The Dynamics Predictor Layer estimates internal contact forces and the accelerations of the inner and outer rings, updating their states through  Euler integration over discrete sub-time steps. Building on these updates, the Kinematics Predictor Layer infers the velocities of the rolling elements by incorporating the latest ring dynamics. Each message-passing step applies   both layers iteratively  to intermediate states, progressively  evolving the system's dynamics over time.

A dedicated  loss function penalizes discrepancies between predicted and ground-truth accelerations and edge forces at both the initial and final sub-time steps,  ensuring accuracy and consistency in dynamic predictions (Section \ref{sec:loss}).

The testing pipeline (Section \ref{sec:test_pipeline}) generates  rollouts iteratively, using  overlapping predictions to reduce  error accumulation and support  robust long-term dynamic modeling.

The following sections provide a detailed description of each component of the framework, beginning  with the graph-based system  representation in Section \ref{sec:graph_rep}.

 \begin{figure}[h!]  
\centering \includegraphics[width=1.0\textwidth]{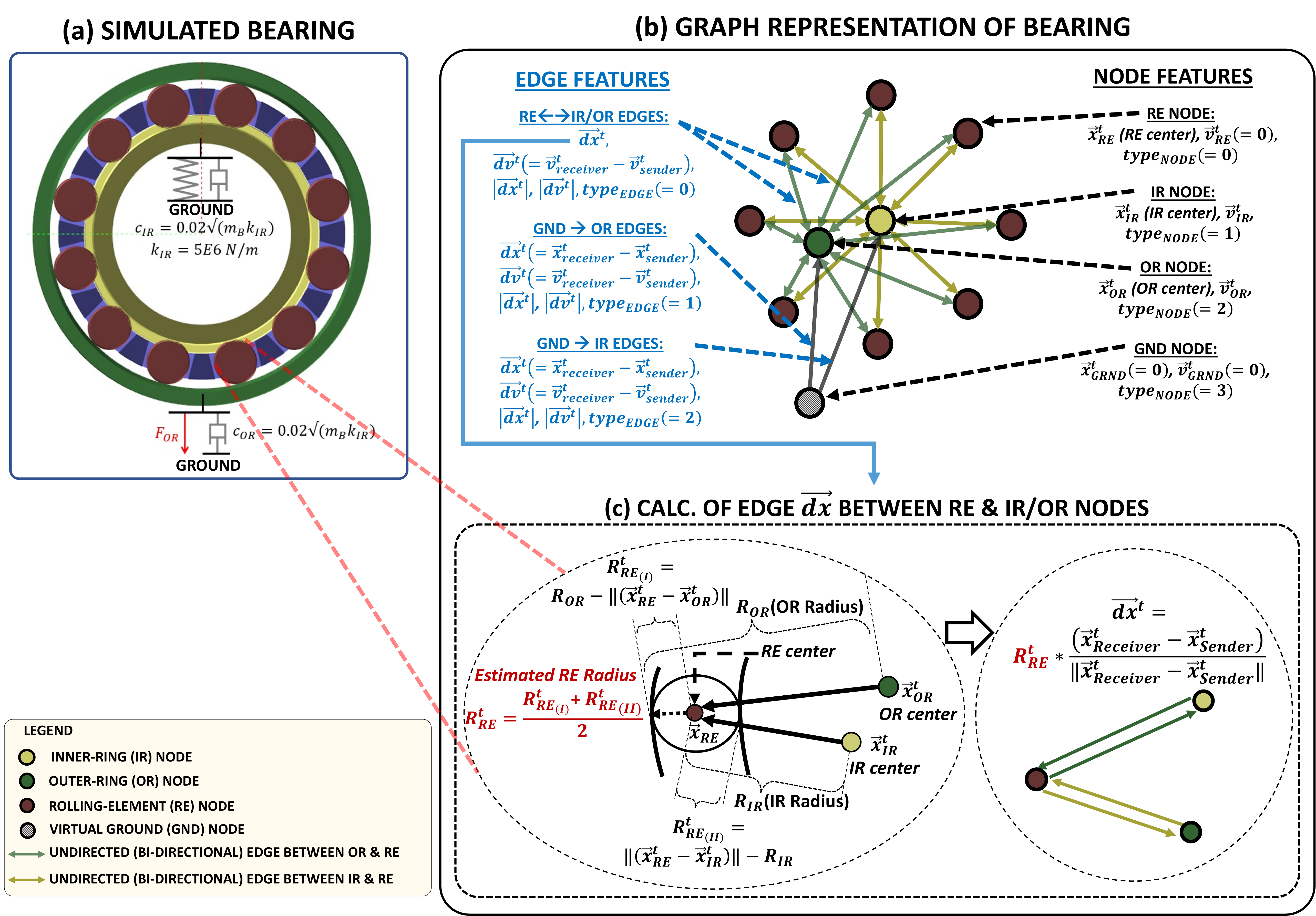}  
\caption{\textbf{Graph Representation of a Ground-Mounted Bearing System}:
\textbf{(a)} Simulated cylindrical roller bearing model showing rollers (brown), with the inner ring (yellow) and outer ring (green) connected to the ground via reaction forces.
\textbf{(b)} Graph representation of the bearing with nodes for rolling elements (RE), inner ring (IR), outer ring (OR), and a virtual ground (GND). Bidirectional edges (RE--IR, RE--OR) capture contact forces between rollers and rings, while directed edges from GND model external reactions on IR and OR.
\textbf{(c)} Computation of the rolling element’s effective radius from IR and OR center distances for the relative position feature, encoding instantaneous roller deformation in contact edges.}
\label{fig:graph_rep}  
\end{figure}   

\subsection{Graph Representation}
\label{sec:graph_rep}
We represent the cylindrical roller bearing as a graph \( G = (V, E) \), where  nodes $V$ correspond to physical components of the bearing, and the edges $E$ encode their pairwise  interactions. This graph-based representation builds on earlier work \cite{sharma2023graph}, which modeled the rolling elements,  inner ring, and  outer ring as nodes to capture contact dynamics. In contrast to the previous approach, we introduce  a virtual ground node that  explicitly models  reaction forces from the ground, improving physical consistency. We also introduce physics-informed edge features that capture  the instantaneous deformed state of the rollers, derived from their relative positions with respect to the inner and outer rings.  Figure \ref{fig:graph_rep} illustrates this representation for a ground-mounted bearing, where the inner ring and outer rings experience ground reaction forces modeled using spring-damper components. The figure highlights how bearing components are mapped onto a structured graph, enabling effective learning of force interactions and state evolution. 

The nodes in the graph are defined as follows:

\begin{equation}  
V = \{ v_{RE_i} \mid i = 1, 2, \dots, n_{RE} \} \cup \{ v_{\text{IR}}, v_{\text{OR}}, v_{\text{GND}} \}  
\end{equation}  

where:  
\begin{itemize}  
    \item \( v_{RE_i} \) for \( i = 1, 2, \dots, n_{RE} \) are the nodes representing the rolling elements.
    \item \( v_{\text{IR}} \) and \( v_{\text{OR}} \) are the nodes representing the inner ring and outer ring respectively.
    \item \( v_{\text{GND}} \) is a virtual node that represents the ground, modeling the bearing housing and ground as a single entity that provides reaction forces and serves as a vibration dissipation sink.  
\end{itemize}  

The edges define the interactions between these components: \begin{equation}  
E = E_{\text{RE-IR}} \cup E_{\text{RE-OR}} \cup E_{\text{GND-IR}} \cup E_{\text{GND-OR}}  
\end{equation}  

where:  
\begin{itemize}
    \item 
    \( E_{\text{RE-IR}} = \Bigl\{\,(v_{RE_i},\,v_{\text{IR}}),\;(v_{\text{IR}},\,v_{RE_i}) \;\bigm|\; i=1,\dots,n_{RE}\Bigr\}\).  
    \\
    (Bidirectional edges between rolling elements nodes \(v_{RE_i}\) and the inner ring node \(v_{\text{IR}}\), modeling contact forces.)
    
    \item 
    \( E_{\text{RE-OR}} = \Bigl\{\,(v_{RE_i},\,v_{\text{OR}}),\;(v_{\text{OR}},\,v_{RE_i}) \;\bigm|\; i=1,\dots,n_{RE}\Bigr\}\).  
    \\
    (Bidirectional edges between rolling elements nodes \(v_{RE_i}\) and the outer ring node \(v_{\text{OR}}\), also modeling contact forces.)
    
    \item 
    \( E_{\text{GND-IR}} = \bigl\{\,(v_{\text{GND}},\,v_{\text{IR}})\bigr\} \).  
    \\
    (A directed edge from the ground node \(v_{\text{GND}}\) to the inner ring node \(v_{\text{IR}}\), modeling external reaction forces.)
    
    \item 
    \( E_{\text{GND-OR}} = \bigl\{\,(v_{\text{GND}},\,v_{\text{OR}})\bigr\} \).  
    \\
    (A separate directed edge from the  ground node \(v_{\text{GND}}\) to the outer ring node \(v_{\text{OR}}\), modeling external reaction forces.)
\end{itemize}
There are no direct edges  connecting different rolling elements or linking the inner ring and outer rings. Instead, only the interactions between the rings and  rolling elements are modeled via bidirectional edges (\(E_{\text{RE-IR}}\), \(E_{\text{RE-OR}}\)), that capture contact forces and enforce Newton’s third law through symmetric force exchange. In contrast, directed edges from the ground node to each ring node (\(E_{\text{GND-IR}}\), \(E_{\text{GND-OR}}\)) represent  external reaction forces, with the ground acting as a fixed reference that applies  forces without requiring  reciprocal predictions. 

To distinguish interaction  types, we assign  a unique  identifier to contact modeling edges (\(E_{\text{RE-IR}} \cup E_{\text{RE-OR}}\)) to reflect their role in modeling  symmetric contact dynamics. Separate  identifiers  are also defined for the ground reaction edges on the inner ring (\(E_{\text{GND-IR}}\)) and outer ring (\(E_{\text{GND-OR}}\)),each  governed  by distinct force functions. This clear separation allows  the model to differentiate  localized contact interactions from external ground reactions, preserving  physical consistency in the representation of a mounted bearing system.

\subsubsection{Node Features}  
\label{sec:node_features}  
Figure \ref{fig:graph_rep}-(b) illustrates the graph representation  corresponding to the bearing configuration  in Figure \ref{fig:graph_rep}-(a). Each node is assigned a set of vector features representing its kinematic state (position and velocity), and scalar features, which serve as node type identifiers that encode structural functions without explicitly modeling inertial properties. 

Each rolling element node \( v_{RE_i} \) for \( i = 1, 2, \dots, n_{RE} \) is  initialized with its center  position \( \vec{x}_{RE_i}^t \) at time \(t\), and a velocity vector \( \vec{v}_{RE_i}^t \) set to zero.  As their motion is constrained by contact  with the inner and outer rings, rolling element  velocities are inferred dynamically through  message passing (Section \ref{sec:model}). A type identifier of 0 is assigned to all rolling element nodes to  distinguish them from other components.

The inner ring node \( v_{\text{IR}} \) and outer ring node \( v_{\text{OR}} \) are initialized with  position vectors \( \vec{x}_{\text{IR}}^t \) and \( \vec{x}_{\text{OR}}^t \), and velocity vectors \( \vec{v}_{\text{IR}}^t \) and \( \vec{v}_{\text{OR}}^t \), at time step \(t\), which evolve over time based on dynamic interactions. The type identifiers for the inner ring and the outer ring nodes are set to 1 and 2, respectively, to differentiate structural properties  and implicitly capture differences  in their inertial behavior.

The ground node \( v_{\text{G}} \) acts  as a fixed reference, with its position and velocity vectors fixed at zero: \( \vec{x}_{\text{G}}^t = 0 \) and \( \vec{v}_{\text{G}}^t = 0 \). It does not evolve  over time and serves solely  as a source of reaction forces and vibration damping . A type identifier of 3 distinguishes it from the physical bearing components and  encodes its constrained   degrees of freedom. A summary of the assigned node features is provided in Table \ref{tab:node_features} .

\begin{table}[h]  
\centering  
\renewcommand{\arraystretch}{1.2} 
\setlength{\tabcolsep}{6pt} 

\begin{tabular}{l cc}  
\hline  
\textbf{Node} & \textbf{Vector Features} & \textbf{Scalar Features} \\  
\hline  
Rolling element (\( v_{RE_i} \)) & \( \vec{x}_{\text{RE}_i}^t, \quad \vec{v}_{\text{RE}_i}^t = 0 \) & \( 0 \) \\  
Inner ring (\( v_{\text{IR}} \)) & \( \vec{x}_{\text{IR}}^t, \quad \vec{v}_{\text{IR}}^t \) & \( 1 \) \\  
Outer ring (\( v_{\text{OR}} \)) & \( \vec{x}_{\text{OR}}^t, \quad \vec{v}_{\text{OR}}^t \) & \( 2 \) \\  
Ground (\( v_{\text{G}} \)) & \( \vec{x}_{\text{GND}}^t = 0, \quad \vec{v}_{\text{GND}}^t = 0 \) & \( 3 \) \\  
\hline  
\end{tabular}  

\caption{Summary of node features in the graph representation of the bearing.}  
\label{tab:node_features}  
\end{table}

\subsubsection{Edge Features}  
\label{sec:edge_features}  
Each edge in the graph is assigned both vector and scalar features. The vector features capture   the relative position and velocity between connected nodes , while the scalar features represent  the magnitudes of these vectors . 

\paragraph{Edge Relative Position Feature}  
Figure \ref{fig:graph_rep}-(c) illustrates the computation of the relative position feature for edges connected to the rolling element nodes. Each bidirectional edge between  a rolling element and the inner ring or outer ring  represents a contact interaction governed by  the deformation of the rolling element. This deformation is characterized by the effective radius of the rolling element, which is computed   from the instantaneous positions of the inner ring, outer ring, and the rolling element.

At a given time \( t \), the effective radius is estimated using the following equations:
\begin{equation}
R_{\text{RE}_{i_{(I)}}}^{t} = ||\vec{x}_{\text{RE}_i}^t - \vec{x}_{\text{IR}}^t|| - R_{\text{IR}}
\end{equation}

\begin{equation}
R_{\text{RE}_{i_{(II)}}}^{t} = R_{\text{OR}} - ||\vec{x}_{\text{RE}_i}^t - \vec{x}_{\text{OR}}^t||
\end{equation}

Here, \( R_{\text{RE}_{i_{(I)}}}^{t} \) represents the estimated radius of a given rolling element based on the distance of its center from the inner ring center, while \( R_{\text{RE}_{i_{(II)}}}^{t} \) provides an alternative estimate using  the distance of its center from the outer ring center. Since the inner ring and the outer ring are predefined structural components, their radii, \( R_{\text{IR}} \) and \( R_{\text{OR}} \), are assumed to be known parameters defined by the  bearing design specifications.

To ensure robustness, the effective radius of the rolling element is computed as the mean of these two estimates:

\begin{equation}
R_{\text{RE}_i}^{t} = \frac{R_{\text{RE}_{i_{(I)}}}^{t} + R_{\text{RE}_{i_{(II)}}}^{t}}{2}
\end{equation}

This effective radius is then used to compute the relative position feature for all edges connected to a rolling element.

\begin{equation}
\Delta \vec{x}_{e}^t = R_{\text{RE}_i}^{t} \cdot \hat{d}_{e}^t, \quad e \in E_{\text{RE-IR}} \cup E_{\text{RE-OR}}
\end{equation}

where \( \hat{d}_{e} \) is the unit edge direction vector, defined as the vector between the sender and receiver nodes of the edge:

\begin{equation}
\hat{d}_{e}^t = \frac{\vec{x}_{\text{receiver}}^t - \vec{x}_{\text{sender}}^t}{||\vec{x}_{\text{receiver}}^t - \vec{x}_{\text{sender}}^t||}
\end{equation}

For directed edges connecting the ground node to the inner ring  (\(E_{\text{GND-IR}}\)) and outer ring (\(E_{\text{GND-OR}}\)), the relative position feature is computed as the difference between the positions of the sender (ground) and receiver (ring) nodes:

\begin{equation}  
\Delta \vec{x}_{e}^t = \vec{x}_{\text{receiver}}^t - \vec{x}_{\text{sender}}^t, \quad e \in E_{\text{GND-IR}} \cup E_{\text{GND-OR}}  
\end{equation} 

\paragraph{Edge Relative Velocity Feature}  
In addition to relative position, each edge includes   a relative velocity feature that captures the difference in velocity between the connected sender and receiver nodes. This feature is essential for modeling dynamic interactions, particularly for learning damping behaviors .

The relative velocity vector for each edge is computed as:

\begin{equation}  
\Delta \vec{v}_{e}^t = \vec{v}_{\text{receiver}}^t - \vec{v}_{\text{sender}}^t, \quad e \in E  
\end{equation}  

By combining relative position and velocity features, the edge representation captures both deformation-induced and dissipative effects, ensuring a physically consistent modeling of force transmission within the bearing system.

\paragraph{Edge Scalar Features}  
Each edge is assigned scalar features that include both interaction-type  identifiers and the magnitudes of the relative position and velocity vectors:

\begin{equation}
\left( |\Delta \vec{x}_e^t|, \quad |\Delta \vec{v}_e^t| \right), \quad e \in E,
\end{equation}
where \( |\Delta \vec{x}_e^t| \) represents the magnitude of the relative position vector, capturing deformation-induced interactions, and \( |\Delta \vec{v}_e^t| \) denotes the magnitude of the relative velocity vector, which encodes damping effects. The edge identifier  \( e \) distinguishes between different interaction types, such as contact forces  between  rolling elements and  rings, and ground reaction forces on the inner  and  outer rings. Table \ref{tab:edge_features} summarizes the edge features for different edges in the bearing graph representation.





\begin{table}[h]
\centering
\renewcommand{\arraystretch}{1.2} 
\setlength{\tabcolsep}{4pt} 
\begin{tabular}{l p{4cm} p{3.5cm}}  
\hline  
\textbf{Edge} & \textbf{Vector Features} & \textbf{Scalar Features} \\  
\hline  
\( e \in E_{\text{RE-IR}}\cup E_{\text{RE-OR}}\) & \( \Delta \vec{x}_e^t(=R_{\text{RE}}^{t} \hat{d}_{e}^t), \, \Delta \vec{v}_e^t \) & \( |\Delta \vec{x}_e^t|, |\Delta \vec{v}_e^t|  \) \\  
\( e \in E_{\text{GND-IR}} \) & \( \Delta \vec{x}_e^t, \, \Delta \vec{v}_e^t \) & \(|\Delta \vec{x}_e^t|, |\Delta \vec{v}_e^t| \) \\
\( e \in E_{\text{GND-OR}} \) & \( \Delta \vec{x}^t_e, \, \Delta\vec{ v}^t_e \) & \(  |\Delta \vec{x}_e^t|, |\Delta \vec{v}_e^t|\) \\  
\hline  
\end{tabular}
\caption{Edge feature definitions for different edge types, where \( \hat{d}_{e} \) is the unit edge direction vector. The edges between rolling elements and rings are bidirectional, while edges from the ground node are directed.}
\label{tab:edge_features}
\end{table}

\paragraph{Feature normalization} 
All vector features associated with nodes and edges are normalized by dividing them by the maximum magnitude observed in the training dataset. In contrast , the  scalar edge features \(|\Delta \vec{x}_e^t|, |\Delta \vec{v}_e^t|\) are scaled using min-max normalization to ensure consistent feature ranges across training samples .

\subsection{Single Equi-Euler GraphNet Layer}
\label{sec:model}
\begin{figure}[h!]  
\centering \includegraphics[width=0.9\textwidth]{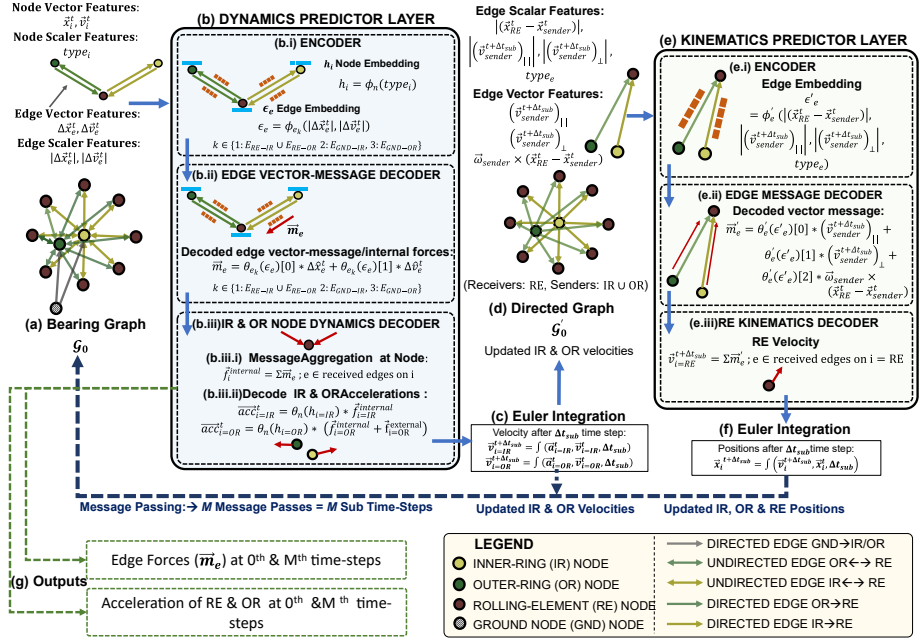}  
\caption{\textbf{Equi-Euler GraphNet: Dynamics and Kinematics Prediction Layers.} The model predicts bearing dynamics by sequentially updating forces, accelerations, and velocities.  
\textbf{(a) Bearing Graph} represents rolling elements (RE), inner ring (IR), outer ring (OR), and ground (GND) as nodes, with bidirectional rolling element–ring edges for contact forces and directed ground–ring edges for external reactions.  
\textbf{(b) Dynamics Predictor Layer} estimates internal forces and ring accelerations. \textbf{(b.i) Encoder} embeds node and edge features. \textbf{(b.ii) Edge Vector-Message Decoder} computes interaction forces from relative distance and velocity vectors. \textbf{(b.iii) Node Dynamics Decoder} aggregates forces, incorporates external reactions, and computes accelerations.  
\textbf{(c) Euler Integration} updates ring node velocities using predicted accelerations over a sub-time step.  
\textbf{(d) Directed Graph} represents the updated state, where directed edges propagate velocity updates from rings to rolling elements.  
\textbf{(e) Kinematics Predictor Layer} estimates rolling element velocities. \textbf{(e.i) Encoder} embeds edge features. \textbf{(e.ii) Edge Vector-Message Decoder} scales velocity components. \textbf{(e.iii) Kinematics Decoder} aggregates edge messages to compute rolling element velocities.  
\textbf{(f) Euler Integration} updates node positions, forming a new system state. Steps (a)–(f) repeat for each message-passing iteration.  
\textbf{(g) Outputs} include interaction forces and accelerations of the rolling elements, inner ring, and outer ring nodes at initial and final sub-time steps.}
\label{fig:model}  
\end{figure}

The Equi-Euler GraphNet  takes as input the graph representation of the bearing (see Figure \ref{fig:model}-(a)), along with the applied external force on the outer ring, the rotational speed in rotations per minute (RPM) of the inner ring, the simulation time step \( \Delta t \), and the hyperparameter specifying the number of message-passing steps \( M \). Each model  layer  processes  the input graph in two sequential stages: (1) the \textbf{dynamics predictor layer} estimates internal contact forces between rolling elements and computes accelerations of the inner ring and outer ring while incorporating external forces, and (2) the \textbf{kinematics predictor layer}, which updates the velocities of the rolling elements based on the inner ring and outer ring dynamics and the prescribed inner ring RPM.

The \textbf{dynamics predictor layer} processes the bearing graph to predict the interaction forces between  rolling element nodes and the ring nodes. These forces are aggregated  to compute  the accelerations of the inner ring and outer rings. The resulting  accelerations are then integrated using a single-step Euler method with a sub-time step \( \Delta t_{\text{sub}} = \Delta t / M \), producing  updated velocities for the inner ring and outer ring nodes.

Next, the \textbf{kinematics predictor layer} updates the velocities of the rolling element nodes. Because  their motion is constrained by contact with the inner and outer rings, their updated velocities are inferred based on the newly computed  velocities of the inner ring and outer ring nodes and  the angular motion of the inner ring. At this stage, the model constructs a directed graph \( \mathcal{G}'_0 \), where edges are directed  only from the inner ring and outer ring nodes to the rolling element nodes. Unlike the dynamics predictor layer,  which uses bidirectional edges  to model contact forces, only these directed edges are assigned features in the kinematics layer. These features include the sender node's velocity projected  along the edge direction and the tangential velocity  induced by angular motion, computed as \( \vec{\omega} \times \Delta \vec{x} \).

The kinematics predictor layer outputs the updated velocities of the rolling element nodes, completing the velocity updates for all nodes in the graph. A final \textbf{Euler integration step} is then applied  using the sub-time step \( \Delta t_{\text{sub}} \) to update the positions of all nodes.

Each full pass through the dynamics and kinematics predictor layers constitutes a single message-passing step, effectively evolving the node states by a sub-time step. The updated graph, with new node positions, velocities, and recalculated edge features, is then fed back into the model. This process is repeated for \( M \) steps to complete the full simulation time step \( \Delta t \).

This section outlines the overall message-passing pipeline. The \textbf{dynamics predictor layer} is detailed in Section \ref{sec:dyn_layer}, followed by the \textbf{kinematics predictor layer} in Section \ref{sec:kin_layer}. 

\subsubsection{Dynamics Predictor Layer}
\label{sec:dyn_layer}

The \textit{Dynamics Predictor Layer} (Figure \ref{fig:model}-(b)) takes as input the bearing graph \( G = (V, E) \), where nodes and edges are assigned scalar and vector features as described in Section \ref{sec:edge_features}. This layer predicts pairwise  internal forces between nodes and computes the accelerations of the inner  and outer ring nodes. 

A key aspect of this layer is its design for  \textit{equivariance} under the Euclidean group E(3)-- in three dimensions-- which includes  translations, rotations, and reflections. This  ensures that all predicted   vector quantities behave consistently with physical laws. Specifically, for  any orthogonal transformation \( R \in \mathbb{R}^{3 \times 3} \) and translation \( b \in \mathbb{R}^3 \), node positions transform as \( x_i \mapsto R x_i + b \), while pairwise  forces \( f_{e} \) and node accelerations \( a_i \) transform equivariantly as \( f_{e}(R x_i + b) \mapsto R f_{e} \) and \( a_i(R x_i + b) \mapsto R a_i \). Scalar features, in contrast, remain invariant. This geometric consistency guarantees that predictions remain  physically plausible and independent of the bearing's orientation or position in 3D space.

\paragraph{Encoder}  
The \textit{Encoder} (Figure \ref{fig:model}-(b.i)) maps scalar node and edge features into high-dimensional embeddings that are invariant under E(3) transformations.

For each node \(v_i \in V\), the scalar attribute \(\text{type}_i\)is passed through a multilayer perceptron (MLP) \(\phi_n\) to produce an E(3)-invariant embedding \(h_i\):
\begin{equation}
    h_i = \phi_n\bigl(\text{type}_i\bigr).
\end{equation}

Each edge \( e \in E \) has the scalar features including  the Euclidean distance \(\|\Delta \vec{x}_{e}^t\|\) and relative velocity magnitude \(\|\Delta \vec{v}_{e}^t\|\) between its connected nodes. These features are inherently \textit{E(3)-invariant}:
\begin{itemize}
    \item Translation invariance arises from using relative quantities  \(\Delta \vec{x}\) and velocity \(\Delta \vec{v}\)
    \item Rotation and reflection invariance is ensured  by using vector magnitudes.
\end{itemize}

These invariant edge features are processed using edge-type-specific multilayer perceptrons (MLPs) \(\phi_{e_k}\), where \( k \) identifies   the type   of physical interaction:

\begin{equation}
    \epsilon_e = \phi_{e_k}\left(\|\Delta \vec{x}_{e}^t\|, \|\Delta \vec{v}_{e}^t\|\right), \quad
    k \in 
    \begin{cases}
        1, & \text{rolling element interactions} \\
        2, & \text{inner ring-ground contacts} \\
        3, & \text{outer ring-ground contacts}
    \end{cases}
\end{equation}

The resulting edge embeddings \(\epsilon_e\) capture  pairwise interactions, while  differentiating  the mechanical roles of  the components involved.

\paragraph{Edge Vector-Message Decoder}  
The \textit{Edge Vector-Message Decoder} (Figure \ref{fig:model}-(b.ii)) generates  edge message vectors that represent pairwise force interactions. For each edge \( e \), the message vector \( \vec{m}_e \) is formed  as a weighted combination  of the unit  direction vectors of relative position \( \Delta \hat{x}_e \) and  relative velocity  \( \Delta \hat{v}_e \). The scalar weights  are decoded from the E(3)-invariant edge embedding \( \epsilon_e \) using edge-type-specific MLPs \( \theta_{e_k} \), where \( k \) indicates the interaction type (e.g., rolling element contact or ground reaction). This design ensures that  \( \vec{m}_e \) remains  E(3)-equivariant: the scalar weights decoded from  \( \epsilon_e \) are E(3)-invariant, while the directional vectors \( \Delta \hat{x}_e \) and   \( \Delta \hat{v}_e \) are inherently  E(3)-equivariant. Each decoder  \( \theta_k \) outputs  two scalar weights that determine the respective contributions of relative position and velocity in constructing the edge message:

\begin{equation}
    \vec{m}_e = \theta_k(\epsilon_e)[0] \cdot \hat{\Delta x}_e + \theta_k(\epsilon_e)[1] \cdot \hat{\Delta v}_e,
    k \in 
    \begin{cases}
        1, & \text{rolling element interactions} \\
        2, & \text{inner ring-ground contacts} \\
        3, & \text{outer ring-ground contacts}
    \end{cases}
\end{equation}

\paragraph{Inner Ring and Outer Ring Dynamics Decoder}  
\label{sec:dyn_decoder}
The \textit{Inner Ring and Outer Ring Dynamics Decoder} (Figure~\ref{fig:model}-(b.iii)) predicts the accelerations of the inner ring and outer ring nodes. First, it calculates the net force vector \(\vec{f}_i^{\text{net}}\) at each ring node $i$ by aggregating the incoming E(3)-equivariant edge messages \(\vec{m}_{e}\) and adding any externally applied force :
\begin{equation}
    \vec{f}_i^{\text{net}} = \sum \vec{m}_{e} \;+\; \vec{f}_i^{\text{external}},
\end{equation}

Next, an MLP \(\theta_n\) decodes a scalar from the invariant node embedding \(h_i\) for each ring node. This scalar is used to scale the net force vector, resulting in an E(3)-equivariant acceleration prediction:

\begin{equation}
    \vec{\text{acc}}_{\text{IR}} = \theta_n\bigl(h_{\text{IR}}\bigr)\;\vec{f}_{\text{IR}}^{\text{net}}, 
    \quad
    \vec{\text{acc}_{\text{OR}}} = \theta_n\bigl(h_{\text{OR}}\bigr)\;\vec{f}_{\text{OR}}^{\text{net}}.
\end{equation}

Since the rolling elements are kinematically constrained by their contact with the rings, their accelerations are not directly computed  in this layer. Instead, their velocities are updated later by the \textit{kinematics prediction layer}.

\subsubsection*{Euler Integration of Inner Ring and Outer Ring Velocities}
\label{sec:euler_int_1}
The velocities of the inner and outer rings are updated using a single-step Euler integration based on their predicted accelerations. The integration  uses a sub-time step defined as:

\begin{equation}
\Delta t_{\text{sub}} = \frac{\Delta T}{M}
\end{equation}
where \( \Delta T \) is the total simulation time step, and \( M \) is the number of message-passing iterations.

Given the predicted accelerations \(\vec{\text{acc}}_{\text{IR}}^t\) and \(\vec{\text{acc}}_{\text{OR}}^t\), the updated velocities at time \( t + \Delta t_{\text{sub}}\) 
  are computed as:

\begin{equation}
\vec{v}_{\text{IR}}^{\,t + \Delta t_{\text{sub}}} 
        = \vec{v}_{\text{IR}}^{\,t} 
           \;+\; \vec{\text{acc}}_{\text{IR}}^t \,\Delta t_{\text{sub}}, 
    \quad
    \vec{v}_{\text{OR}}^{\,t + \Delta t_{\text{sub}}} 
        = \vec{v}_{\text{OR}}^{\,t} 
           \;+\; \vec{\text{acc}}_{\text{IR}}^t \,\Delta t_{\text{sub}}
    \label{eq:IR,ORvt+1}
\end{equation}

These updated velocities of the inner  and outer ring nodes are subsequently  used in the  \textit{Kinematics Predictor Sub-Layer} to compute the resulting motion of the rolling element nodes.

\subsubsection{Kinematics Predictor Layer}  
\label{sec:kin_layer}  
The \textit{Kinematics Predictor Layer} (Figure~\ref{fig:model}-(e)) updates the velocities of the rolling element nodes based on the predicted motion of the inner  and outer ring nodes. It operates on a directed graph \( G'=(V',E')\) (Figure~\ref{fig:model}-(d)), where the node set \(V' = \{ v_{\text{IR}}, v_{\text{OR}}, v_{\text{RE}_i} \}\) represents the bearing components, and the edge set \(E' = E'_{\text{IR-RE}} \cup E'_{\text{OR-RE}}\) consists of directed edges from the inner and outer rings to each rolling element, representing  motion transfer pathways.

As with the \textit{Dynamics Predictor Layer}, all transformations within the \textit{Kinematics Predictor Layer} are explicitly designed to be E(3)-equivariant, ensuring that the predicted rolling element velocities  inherently preserve consistency under translation, rotation, and reflection. 

However, unlike the \textit{Dynamics Predictor Layer} -- which uses node-centric kinematic states -- the \textit{Kinematics Predictor} encodes  all relevant information as scalar and vector features on edges. This design reflects the inherently directed nature  of motion propagation  from  the rings to the rolling elements and enables physically consistent velocity updates across the system.


For each edge \( e \in E' \), vector features capture the relative kinematics between the sender node (inner  or outer ring) and the receiver  node (rolling element). These features 
 include:
\begin{itemize}
    \item Relative distance vector: 
    \[
    \Delta\vec{x}_e^t = \vec{x}_{\text{receiver}}^t - \vec{x}_{\text{sender}}^t,
    \]
    which describes the spatial offset  between connected nodes.
    \item Decomposed sender velocity:
    \begin{itemize}
        \item Parallel component, \( \vec{v}_{\text{sender} \parallel e}^{t + \Delta t_{\text{sub}}} \), representing  motion along the edge direction.
        \item Perpendicular component, \( \vec{v}_{\text{sender} \perp e}^{t + \Delta t_{\text{sub}}} \), capturing  motion orthogonal to the edge.
    \end{itemize}
    \item Tangential velocity  induced by  rotation:
    \[
    \vec{\omega}_{\text{sender}}^t \times \Delta\vec{x}_e^t,
    \]
    which represents the rotational contribution  to the receiver's  (rolling element’s) velocity.
\end{itemize}

In addition to vector attributes, each edge is assigned scalar features, including the magnitudes of the relative position and velocity vectors, as well as a type identifier \( \text{type}_e \), which distinguishes  between inner ring-rolling element and outer ring-rolling element interactions. 
The edge type is defined as:
\begin{equation}
\text{type}_e =
\begin{cases}
0, & e \in E'_{\text{IR-RE}} \\
1, & e \in E'_{\text{OR-RE}}
\end{cases}
\end{equation}
By structuring the Kinematics Predictor Layer in this way, the model effectively propagates velocity updates from the inner  and outer ring nodes to the rolling element nodes, ensuring  a physically consistent representation of motion transfer throughout  the bearing system.

\paragraph{Encoder}  
The Encoder (Figure~\ref{fig:model}-(e.i)) maps  the scalar edge  features to high-dimensional, E(3)-invariant edge embeddings \( \epsilon'_{e} \) using an MLP \( \phi_{e'} \):

\begin{equation}
    \epsilon'_{e} = \phi_{e'} \Big( \|\Delta \vec{x}_{e}^t\|, \;\|\vec{v}_{\text{sender} \parallel e}^{t + \Delta t_{\text{sub}}}\|, \;\|\vec{v}_{\text{sender} \perp e}^{t + \Delta t_{\text{sub}}}\|, \;\|\vec{\omega}_{\text{sender}}^{t} \times \vec{\Delta x}_e^t\|, \;\text{type}_e \Big).
\end{equation}

\paragraph{Edge Vector-Message Decoder}  
The \textit{Edge Vector-Message Decoder} (Figure~\ref{fig:model}-(e.ii)) generates velocity messages along edges to propagate motion information toward  the rolling element nodes. For each directed edge \( e \), a velocity message is constructed by scaling the edge vector features with learned scalar weights. These weights are obtained by decoding the E(3)-invariant edge embedding \( \epsilon'_{e} \) using an MLP \( \theta_{e'} \). This architecture is explicitly designed to  preserve E(3)-equivariance: the scalar weights remain invariant  under translations, rotations, and reflections  due to their dependence on the invariant embedding, while the edge vector features  transform equivariantly, ensuring that the resulting velocity messages behave consistently under Euclidean transformations: 

\begin{align}
    \vec{m}_{e}^{'} &= \theta_{e'}(\epsilon'_{e})[0] \cdot \vec{v}_{\text{sender} \parallel e }^{t + \Delta t_{\text{sub}}} 
    + \theta_{e'}(\epsilon'_{e})[1] \cdot \vec{v}_{\text{sender} \perp e }^{t + \Delta t_{\text{sub}}} \nonumber \\
    &\quad + \theta_{e'}(\epsilon'_{e})[2] \cdot \big(\vec{\omega}_{\text{sender}}^t \times \vec{\Delta x}_{e}^t\big).
\end{align}
These edge messages encode  the velocity contributions from  the inner  and outer ring nodes to their corresponding rolling element nodes.

\paragraph{Rolling Element Kinematics Decoder}  
The \textit{Rolling Element Kinematics Decoder} (Figure \ref{fig:model}-(e.iii)) aggregates  the incoming E(3)-equivariant velocity messages at each rolling element node to compute its updated velocity. This aggregation captures the cumulative effect of surrounding ring dynamics on each rolling element, while preserving the geometric consistency of the motion.

\begin{equation}
    \vec{v}^{t + \Delta t_{\text{sub}}}_i = \sum \vec{m}_{e}^{'}, \quad i \in \text{RE}
    \label{eq:REvt+1}
\end{equation}

This ensures that the velocity of each rolling element node is computed  based on contributions from both the inner and outer ring nodes, thereby preserving  the kinematic constraints inherent to the bearing system.

\subsubsection*{Euler Integration for Updated Positions}
\label{sec:euler_int_2}
The updated velocities of the rolling element nodes (Equation \ref{eq:REvt+1}), along with those of the inner and outer ring nodes (Equation \ref{eq:IR,ORvt+1}), are integrated over the sub-time step to obtain  their new positions:

\begin{equation}
    \vec{x}^{t+\Delta t_{\text{sub}}}_i = \vec{x}^{t}_i + \frac{(\vec{v}^{t}_i+\vec{v}^{t + \Delta t_{\text{sub}}}_i)}{2} \cdot \Delta t_{\text{sub}}, \quad i \in \text{RE}\cup\text{IR}\cup\text{OR}.
\end{equation}
This final integration step updates the state of all nodes, completing one full message-passing iteration. The resulting  graph, with updated node positions, velocities, and recomputed  edge features, is then passed  back into the model for the next iteration. This process repeats until the specified number of message-passing steps is reached.

\subsection{Loss Formulation}
\label{sec:loss}

The final outputs of the \textit{Dynamics Predictor Layer} include predicted accelerations for the inner and outer ring nodes, as well as edge-level contact forces, at both the initial  (\(t_0\)) and a final  (\(t_N\)) sub-time steps within each simulation time step. Let
\[
\vec{\text{acc}}_{i}^{t_0}, \;\vec{\text{acc}}_{i}^{t_N} 
\quad \text{and} \quad 
\vec{\text{acc}}_{i}^{M_0}, \;\vec{\text{acc}}_{i}^{M_N}
\]
denote the ground-truth and model-predicted accelerations for node \(i \in \{\text{IR}, \text{OR}\}\) (inner and outer rings) at these two sub-time steps. Similarly, let
\[
\vec{f}_{e}^{t_0}, \;\vec{f}_{e}^{t_N} 
\quad \text{and} \quad 
\vec{f}_{e}^{M_0}, \;\vec{f}_{e}^{M_N}
\]
represent the corresponding ground-truth and predicted contact forces on each edge \(e \in E\), where $E$ is the set of  edges in the graph-based representation of the bearing.

To promote temporal  stability, we penalize prediction errors at both the initial and final sub-time steps. The \emph{initial-step loss} measures  the model's  accuracy when predicting accelerations and forces from the ground-truth state, while the \emph{final-step loss} regularizes the model under autoregressive conditions---i.e., when it uses its own previous predictions as input. This dual-loss strategy helps mitigate \emph{distribution shift} in the input data, which can lead to compounding errors over long simulation rollouts.

Formally, we define the total loss as:
\[
\mathcal{L}_{\text{total}} \;=\; 
\lambda_a \Bigl(\,\mathcal{L}_{\text{acc}}^{t_0} \;+\;\mathcal{L}_{\text{acc}}^{t_N}\Bigr) 
\;+\; 
\lambda_f \Bigl(\,\mathcal{L}_{\text{force}}^{t_0} \;+\;\mathcal{L}_{\text{force}}^{t_N}\Bigr),
\]
where \(\mathcal{L}_{\text{acc}}^{t_k}\) and \(\mathcal{L}_{\text{force}}^{t_k}\) are the mean-squared errors (MSE) in acceleration and contact force predictions at sub-time steps \(t_k \in \{t_0,t_N\}\). The scalar coefficients  \(\lambda_a\) and \(\lambda_f\) control the relative importance  of acceleration versus  force accuracy.

This \emph{two-step} loss formulation is inspired by the \emph{pushforward trick}~\cite{brandstetter2022message}, which addresses a key challenge in learned dynamical simulators: the shift from training on ground-truth inputs to deploying auto-regression on self-generated predictions. By enforcing prediction consistency at both the initial and  final sub-time step---particularly at the final step, where the model has already propagated its own outputs---we encourage  robustness against  compounding errors. In essence, the model learns to self-correct, reducing  reliance on perfect inputs and improving stability in long-horizon rollouts.

\subsection{Testing Pipeline}
\label{sec:test_pipeline}

During testing, as illustrated in Figure~\ref{fig:overview}-(c), the model begins from the known initial bearing state at \(t=0\) and generates predictions for both \(t=0\) and \(t=N\). In the subsequent  iteration, it uses the predicted state at \(t=N\) as input and produces predictions for \(t=N\) and \(t+2N\). Since the state at \(t=N\) is predicted in both iterations, these overlapping predictions are averaged to obtain  a smoothed and more reliable estimate. This process continues -- each iteration producing overlapping predictions that are averaged -- until  the desired final time  is reached. By chaining together these overlapping intervals of length \(N\), the model constructs a stable and consistent  trajectory,  effectively reducing noise and mitigating the accumulation of rollout errors.

\section{Case Study}
\label{sec:case_study}
\subsection{Problem Setup}
\begin{figure}[h!]  
\centering \includegraphics[width=0.9\textwidth]{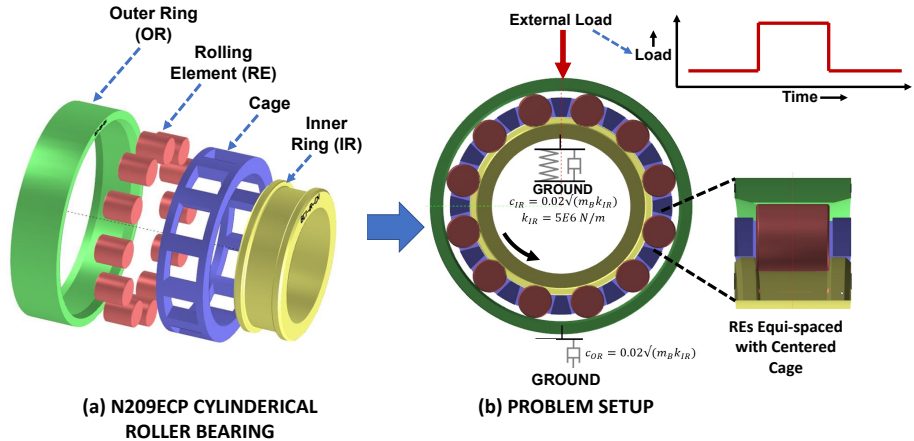}  
\caption{\textbf{Problem Setup for Cylindrical Roller Bearing Dynamics}:
\textbf{(a)} Exploded diagram of the simulated N209ECP bearing assembly with inner ring, outer ring, rolling elements, and a simple cage.
\textbf{(b)} Schematic of the 2D, three-degree-of-freedom multi-body simulation in BEAST, where the outer ring is fixed, an external step load is applied at the 12~o’clock position, doubled at the 2500\textsuperscript{th} time step, and subsequently relaxed at the 5000\textsuperscript{th} time step --- inducing transient vibrations --- while the inner ring rotates at a prescribed speed. Springs and dampers connect the bearing to ground, modeling reaction forces. Rolling elements, centered by a simple cage, respond freely to these transient load variations. The resulting dataset serves as the training and testing basis for the Equi-Euler GraphNet framework.}
\label{fig:problem_setup}  
\end{figure}
In this section, we present a detailed overview of the simulation setup used  to train and evaluate  the proposed Equi-Euler GraphNet framework for modeling the dynamics of cylindrical roller bearings. The training and testing datasets were generated  using the BEAring Simulation Tool (BEAST), a high-fidelity multiphysics simulator developed by SKF  that combines FEA with elastohydrodynamic lubrication modeling. Multiple bearing configurations were simulated, each subjected to an external step load applied to the outer ring. This loading induces transient vibrations that propagate through the bearing components, affecting both load distribution and dynamic response. Concurrently, the inner ring rotates at a predefined  constant  speed. Figure \ref{fig:problem_setup} illustrates the simulated 
 cylindrical roller bearing, highlighting the key components and configuration details relevant to the study.

 \paragraph{Train and Test Dataset}
The training dataset consisted of  6000-step simulated trajectories with a uniform time step of \(6.667 \times 10^{-5} \text{ s}\), generated  across a range of roller counts, rotational speeds, and external loads. Specifically, the configurations included bearings with 12, 13, and 14 rollers, operating at rotational speeds of 0, 300, 333, 375, 429, and 750 RPM under radial  loads of 5, 7, 9, 11, 13, 15, and 17 kN. To ensure physically meaningful training data, simulations were  recorded only after the inner ring reached its  target rotational speed and the initial acceleration-induced transients had dissipated. Early-stage  non-physical forces  resulting from rapid acceleration  were excluded  to prevent the model from learning unrealistic or spurious behavior. After 2500 recorded steps, the applied 
 load was suddenly doubled, inducing  a strong transient response  across all bearing components. At step 5000, the load was abruptly reduced back  to its original  value, triggering a second transient phase. This dataset captures a wide range of dynamic behaviors,  enabling the model to learn the system’s response to changing operational and loading conditions.

The testing data was designed to evaluate the model's performance in both interpolation and extrapolation scenarios. For the interpolation case, a bearing with 13 rollers was tested under a 13 kN load at 600 RPM—conditions where both the roller count and load were within the training distribution, but the rotational speed (600 RPM) had not been encountered during training. To assess the model’s out-of-distribution extrapolation ability, we tested the model in two progressively challenging scenarios. In the first scenario, a test case  with a 13-roller bearing operating at a 15 kN load and 1500 RPM was used, introducing a rotational speed beyond the training range. In the second scenario, an extreme extrapolation scenario was considered: a bearing with 15 rollers subjected to a 19 kN load at 3000 RPM, representing conditions far outside the training distribution. A comprehensive summary of all training and test configurations is provided in Table~\ref{tab:train_test_conditions}.

\begin{table}[h!]
\centering
\caption{Operating conditions used in training and test evaluations.}
\scriptsize
\renewcommand{\arraystretch}{1.3}
\setlength{\tabcolsep}{6pt}
\begin{tabular}{p{3.2cm}|p{2.3cm}|p{1.5cm}p{1.8cm}p{1.8cm}}
\hline
\textbf{Operating Condition} & \textbf{Train} & \multicolumn{3}{c}{\textbf{Testing Scenarios}} \\
\cline{3-5}
& & \textbf{Interp.} & \textbf{Extrap. 1} & \textbf{Extrap. 2} \\
\hline
Number of Rollers         & 12, 13, 14 & 13 & 13 & 15 \\
Load (kN)                 & 5, 7, 9, 11, 13, 15, 17 & 13 & 15 & 19 \\
\shortstack[l]{Rotational Speed (RPM)} & 0, 300, 333, 375, 429, 750 & 600 & 1500 & 3000 \\
\hline
\end{tabular}
\label{tab:train_test_conditions}
\end{table}


\subsection{BEAST: Advanced Bearing Simulator}
\label{sec:beast}
BEAST simultaneously solves the coupled dynamics of multi-body systems, structural deformation, thermal effects, and lubricated contact conditions. It incorporates elastohydrodynamic lubrication, tribological effects, and complex surface geometries to accurately capture complex contact interactions. While all bearing components are modeled as rigid bodies, elastic deformation  within the contact zones is accounted for using a dedicated contact solver. To handle the varying dynamics, BEAST employs an implicit predictor-corrector scheme with adaptive  time stepping -- smaller time  steps  are used to resolve highly nonlinear contact events, while larger steps improve computational efficiency under  near-linear conditions. This adaptive  integration strategy  enables accurate simulation  of both transient and steady-state behaviors in rolling bearings.

In this study,  the bearing system is modeled  as a constrained two-dimensional representation by assuming perfect alignment bearing and a purely radial load, with the cage guided solely by the outer ring. Under these simplified yet representative conditions, the cylindrical roller bearing operates  as a two-dimensional multi-body system with three degrees of freedom (3-DOF): two translational and one rotational. This formulation captures the in-plane dynamics of the system while intentionally  neglecting out-of-plane effects to reduce complexity without compromising the fidelity of key dynamic behaviors.

The simulation setup is illustrated in Figure \ref{fig:problem_setup}. To enforce the applied rotational speed and eliminate  misalignment, the inner ring was fixed to a rotating coordinate system. Additionally,  it was connected to the ground using a weak spring (\(k=5 \times 10^6 N/m\)) and a damper set to \(1 \%\) of the critical damping.

The outer ring was fixed in a stationary coordinate system to prevent rotational motion and was damped at \(1 \%\) of the critical damping. A vertical external step load was applied at the 12 o’clock position -- first  applied abruptly, then doubled in magnitude, and subsequently released. This   loading profile created a clearly defined loaded zone and triggered  transient vibrations, which are essential  for analyzing the bearing’s dynamic response under varying load conditions.

The rolling elements were free  to rotate about their own axes and translate  within the bearing plane, while  axial motion, tilt, and skew were restricted to preserve the two-dimensional nature  of the simulation. To maintain uniform spacing between  rollers, a simple cage mechanism was introduced. The cage was constrained in all degrees of freedom except  rotation about the bearing axis, allowing  it to rotate freely while ensuring consistent roller positioning throughout the simulation.

Each BEAST simulation for this configuration required approximately two hours of computation time on a single processor. Simulation outputs were sampled with a uniform time step of \(6.67\times10^{-5} \text{sec}\), and a low-pass filter was applied  to remove high-frequency fluctuations caused  by roller impacts in the unloaded zone. This preprocessing step  ensured a clean and consistent  dataset for robust evaluation of the proposed framework.

\section{ GNN Architectures Used for Benchmarking}
\label{sec:comp_soa}

\begin{figure}[h!]  
\centering \includegraphics[width=0.9\textwidth]{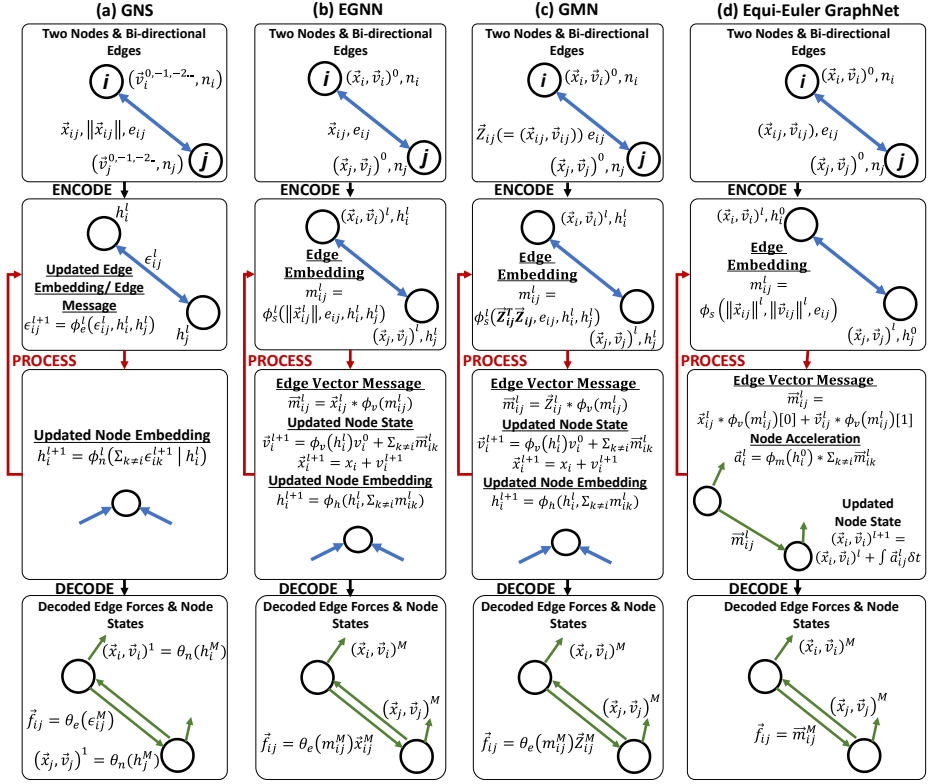}  
\caption{\textbf{Comparison of GNN Architectures for Bearing Dynamics}:
\textbf{(a)--(d)} Schematic of four message-passing frameworks (GNS\cite{sanchez2020learning}, EGNN\cite{satorras2021n}, GMN\cite{huang2022equivariant}, and Equi-Euler GraphNet). While all models follow an encode--process–-decode paradigm, they differ in their feature representations, message construction, and state update mechanisms. GNS (\textbf{a}) encodes stacked scalar and vector features into latent node and edge embeddings, which are iteratively updated and finally decoded as node accelerations and pairwise forces. EGNN (\textbf{b}) and GMN (\textbf{c}) construct invariant scalar edge embeddings, transform them into vector messages, and update node embeddings and states through equivariant message passing. Since these models do not natively predict forces, we append equivariant decoders to map final edge embeddings to interaction forces. In contrast, Equi-Euler GraphNet (\textbf{d}) directly interprets edge messages as physical forces and updates node positions and velocities via Euler integration, enabling simultaneous prediction of trajectories and forces.}
\label{fig:model_comparison}  
\end{figure}

We compare the proposed Equi-Euler GraphNet with three established GNN-based architectures widely used for dynamics prediction: Graph Neural Simulator \(\mathrm{(GNS)}\)\cite{sanchez2020learning}, Equivariant Graph Neural Network \(\mathrm{(EGNN)}\)\cite{satorras2021n}, and Graph Mechanics Network \(\mathrm{(GMN)}\)\cite{huang2022equivariant}. All models adopt an encode-process-decode framework but differ significantly  in how they represent physical quantities, construct messages, and model  system dynamics (See Figure \ref{fig:model_comparison}). GNS treats features distinctly from the rest of the models. It stacks scalar and vector features into latent embeddings without explicitly separating them as done by EGNN, GMN, and Equi-Euler GraphNet. It predicts both node accelerations and pairwise forces from these learned representations. In contrast, EGNN and GMN explicitly separate scalar and vector quantities, using E(3)-equivariant message passing to preserve geometric consistency under Euclidean transformations. However, neither EGNN nor GMN was originally designed to predict forces. To enable a fair comparison,  we extend both  with additional equivariant decoders that map learned edge embeddings to force vectors. The proposed Equi-Euler GraphNet adopts a different approach. 
Rather than relying on abstract latent representations, it interprets edge vectors directly as physical interaction forces and integrates system states using these forces. This design allows it to natively and simultaneously predict both trajectory rollouts  and pairwise interactions. 

\paragraph{GNS (Figure~\ref{fig:model_comparison}(a))}
GNS does not explicitly distinguish between scalar and vector features. Instead, it stacks the past velocity vectors \(\{\vec{\mathbf{v}}_i^{(0)},\dots,\vec{\mathbf{v}}_i^{(5)}\}\) along with any scalar node features into a single node feature vector. Similarly, edge features are constructed by concatenating the relative position vector \(\vec{\mathbf{x}}_{ij}\), its magnitude \(\|\vec{\mathbf{x}}_{ij}\|\), and scalar edge features \(e_{ij}\) into an edge feature vector. These concatenated  features are then encoded with MLPs into node embeddings \(h_i^l\) and edge embeddings \(\epsilon_{ij}^l\). At each message-passing layer \(l\), the edge embeddings are transformed into updated embeddings \(\epsilon_{ij}^l+1\) and used to construct messages \(\mathbf{m}_{ij}^l\), which are aggregated to update the node embeddings \(h_i^{l+1}\) as follows:
\begin{equation}
    \mathbf{m}_{ij}^{l} \;=\; \phi_m\bigl(\epsilon_{ij}^{l},\,h_{i}^{l},\,h_{j}^{l}\bigr),
    \quad
    h_{i}^{l+1} 
    \;=\; 
    \phi_h\!\Bigl(\sum_{k \neq i}\mathbf{m}_{ik}^{l},\,h_{i}^{l}\Bigr).
\end{equation}
where \(\phi_m\) and \(\phi_h\) are MLPs. In subsequent message-passing layers, the updated node and edge embeddings are iteratively  transformed to compute new messages, allowing both representations to evolve jointly across layers. After \(M\) such layers, the final node embeddings \(h_i^M\) are decoded using an MLP \(\theta_n\) to predict node accelerations \(\vec{\mathbf{a}}_i\), while  the final edge embeddings \(\epsilon_{ij}^M\) are decoded using MLP \(\theta_e\) to estimate pairwise interaction forces \(\vec{\mathbf{f}}_{ij}\):
\begin{equation}
    \vec{\mathbf{a}}_i = \theta_n(h_i^M), \qquad
    \vec{\mathbf{f}}_{ij} = \theta_e(\epsilon_{ij}^M).
\end{equation}
The predicted accelerations are then  explicitly integrated to update the positions \(\vec{\mathbf{x}}_i\) and velocities \(\vec{\mathbf{v}}_i\), resulting in a simulation framework that is primarily driven by learned acceleration dynamics.

\paragraph{EGNN and GMN (Figure~\ref{fig:model_comparison}(b)--(c))}  
EGNN and GMN, in contrast to  GNS, explicitly distinguish between scalar and vector-valued features and employ  equivariant message-passing mechanisms that  preserve geometric consistency  under Euclidean transformations.  At each message-passing layer \(l\), scalar node attributes are encoded as \(h_i^l\), while positions \(\vec{\mathbf{x}}_i^l\) and velocities \(\vec{\mathbf{v}}_i^l\) are treated  as vector-valued quantities.

To construct edge messages, a scalar edge embedding \(\mathbf{m}_{ij}^{l}\) is first computed using the relative position between nodes \(i\) and \(j\):
\begin{equation}
    \mathbf{m}_{ij}^{l} 
    \;=\; 
    \phi_s\bigl(\|\vec{\mathbf{x}}_{ij}^l\|,\,e_{ij},\,h_i^l,\,h_j^l\bigr)
\end{equation}

where, \(\vec{\mathbf{x}}_{ij}^l \;=\; \vec{\mathbf{x}}_i^l -    \vec{\mathbf{x}}_j^l\) denotes the relative position vector between node \(i\) and node \(j\) at message-passing layer \(l\), and \(\|\vec{\mathbf{x}}_{ij}^l\|\) is the corresponding  Euclidean distance. The term \(e_{ij}\) represents    scalar edge features, while \(h_i^l\) and \(h_j^l\) are the scalar node embeddings of nodes \(i\) and \(j\), respectively. The function \(\phi_s\) is an MLP that processes these inputs to learn a scalar edge embedding \(\mathbf{m}_{ij}^l\)

\emph{EGNN}  constructs a directional vector message \(\vec{\mathbf{m}}_{ij}^{l}\) by scaling the relative position vector with a learned scalar coefficient derived from the edge embedding :
\begin{equation}
    \vec{\mathbf{m}}_{ij}^{l}
    \;=\;
    \vec{\mathbf{x}}_{ij}^l \cdot \phi_v\bigl(\mathbf{m}_{ij}^{l}\bigr),
\end{equation}
where \(\phi_v\) is an MLP that outputs a scalar weight conditioned   on the scalar  edge embedding \(\mathbf{m}_{ij}^{l}\). This formulation guarantees  that the resulting   message is equivariant with respect to Euclidean transformations, such 
 as rotations and translations, preserving the geometric consistency of the dynamics .

\emph{GMN} extends the EGNN formulation by incorporating both relative positions and relative velocities into a unified, multi-channel edge feature vector. It first computes the relative velocity between nodes \(i\) and \(j\) as \(\vec{\mathbf{v}}_{ij}^l = \vec{\mathbf{v}}_i^l - \vec{\mathbf{v}}_j^l\), and concatenates it with the relative position vector \(\vec{\mathbf{x}}_{ij}^l\) to form the edge feature vector:
\[
    \mathbf{Z}_{ij}^l = [\vec{\mathbf{x}}_{ij}^l,\, \vec{\mathbf{v}}_{ij}^l].
\]
where \(\mathbf{Z}_{ij}^l\) is the multi-channel edge feature vector at message passing layer \(l\).  GMN then computes a scalar edge embedding and a directional vector message as follows:
\begin{equation}
    \mathbf{m}_{ij}^{l}
    \;=\;
    \phi_s\bigl(\mathbf{Z}_{ij}^{l\,\top} \mathbf{Z}_{ij}^l,\, e_{ij},\, h_i^l,\, h_j^l\bigr), \qquad
    \vec{\mathbf{m}}_{ij}^{l}
    \;=\;
    \mathbf{Z}_{ij}^l \cdot \phi_v\bigl(\mathbf{m}_{ij}^{l}\bigr),
\end{equation}

where \(\phi_s\) and \(\phi_v\) are MLPs. The term \(\mathbf{Z}_{ij}^{l\,\top} \mathbf{Z}_{ij}^l\) is the inner product of the multi-channel edge vector, \(e_{ij}\) denotes   scalar edge features, and \(h_i^l, h_j^l\) are the scalar node embeddings for nodes \(i\) and \(j\) at layer \(l\).

Both EGNN and GMN aggregate the directional messages \(\vec{\mathbf{m}}_{ij}^{l}\) at each node to update its velocity, position, and scalar embedding for the next layer \(l+1\). The updates are computed as follows :
\begin{equation}
    \vec{\mathbf{v}}_i^{l+1}
    \;=\;
    \phi_v(h_i^l) \cdot \vec{\mathbf{v}}_i^0
    +
    \sum_{k \neq i} \vec{\mathbf{m}}_{ik}^l, \quad
    \vec{\mathbf{x}}_i^{l+1}
    \;=\;
    \vec{\mathbf{x}}_i^l
    +
    \vec{\mathbf{v}}_i^{l+1}, \quad
    h_i^{l+1}
    \;=\;
    \phi_h\left(h_i^l,\, \sum_{k \neq i} \vec{\mathbf{m}}_{ik}^l\right)
\end{equation}
where \(\vec{\mathbf{v}}_i^{l+1}\) is the updated velocity of node \(i\), computed by combining the scaled initial velocity \(\vec{\mathbf{v}}_i^0\) with the sum of incoming directional messages. The scaling factor  \(\phi_v(h_i^l)\), implemented as an MLP,  is learned  based on the node’s   scalar embedding \(h_i^l\), allowing the model to modulate the influence of the initial velocity. The new position \(\vec{\mathbf{x}}_i^{l+1}\) is obtained by integrating the updated velocity, while the scalar embedding \(h_i^{l+1}\) is updated using another MLP \(\phi_h\), which takes as input  the current embedding and 
 the aggregated directional   messages.

Since neither EGNN nor GMN is originally designed to predict   interaction forces, we extend both models by incorporating an additional decoder MLP \(\theta_e\) that maps the final edge embeddings to force vectors . In EGNN, the force vector \(\vec{\mathbf{f}}_{ij}\) is  computed by scaling the final relative position vector  \(\mathbf{x}_{ij}^M\) 
with a learned scalar output from the decoder:
\begin{equation}
    \vec{\mathbf{f}}_{ij} = \theta_e(\mathbf{m}_{ij}^M) \cdot \vec{\mathbf{x}}_{ij}^M.
\end{equation}
In GMN, the predicted force instead leverages  the full multi-channel  edge representation \(\mathbf{Z}_{ij}^M\), incorporating both relative position and velocity:
\begin{equation}
    \vec{\mathbf{f}}_{ij} = \theta_e(\mathbf{m}_{ij}^M) \cdot \mathbf{Z}_{ij}^M.
\end{equation}

This extension enables both models to produce interpretable pairwise force estimates, facilitating direct comparison with physically meaningful ground-truth interaction forces.

\paragraph{Equi-Euler GraphNet (Figure~\ref{fig:model_comparison}(d))}
Equi-Euler GraphNet also adopts  an equivariant  message-passing strategy, but with a key distinction: it directly  interprets each edge message vector as a physical interaction force, computed solely from edge-level  features.  At each layer, the model first computes a scalar edge embedding as:
\begin{equation}
    \mathbf{m}_{ij}^{l}
    \;=\;
    \phi_s\bigl(\|\vec{\mathbf{x}}_{ij}\|,\|\vec{\mathbf{v}}_{ij}\|,\,e_{ij}\bigr),
\end{equation}
where \(\vec{\mathbf{x}}_{ij}\) and \(\vec{\mathbf{v}}_{ij}\)
  denote the relative position and velocity vectors, respectively, and \(e_{ij}\)
represents the scalar edge features.
 This scalar embedding is then used to construct the vector-valued message:
\begin{equation}
    \vec{\mathbf{m}}_{ij}^{l}
    \;=\;
    \vec{\mathbf{x}}_{ij}\,\phi_v\bigl(\mathbf{m}_{ij}^{l}\bigr)[0]
    \;+\;
    \vec{\mathbf{v}}_{ij}\,\phi_v\bigl(\mathbf{m}_{ij}^{l}\bigr)[1],
\end{equation}
which is directly interpreted as the force vector acting from node $j$ on node $i$. The net force on node $i$ is obtained by summing  the incoming messages \(\vec{\mathbf{m}}_{ij}^{l}\) across all connected neighbors. . This net force is then scaled by a constant node-specific embedding \(\mathbf{h}_i^{0}\) to compute the acceleration.

Rather than relying on iterative updates to node embeddings, Equi-Euler GraphNet uses a fixed feature representation and applies a simple Euler integration to update positions \(\vec{\mathbf{x}}_i\) and \(\vec{\mathbf{v}}_i\) at each layer. By the final iteration, these edge vectors  \(\vec{\mathbf{m}}_{ij}^{l}\) 
 naturally represent the predicted pairwise interaction  forces, enabling simultaneous trajectory rollout  and interaction-force inference  in a physically grounded and interpretable manner.

\paragraph{Adaptation of the Models for Fair Comparison and Experimental Outcomes.}
To ensure a fair comparison, all models are provided with the same graph representation described  in Section \ref{sec:graph_rep}. However, GNS receives  additional input in the form of the past five velocity  vectors for  each bearing component, consistent with its original design \cite{sanchez2020learning}. All GNS features are standardized using a z-score normalization (standard scaler), as originally proposed. In contrast, EGNN and GMN use the same node  scalar and vector features as Equi-Euler GraphNet. Vector features are scaled by their maximum magnitude, while scalar edge features  are normalized using min–max scaling -- again, following the practices outlined in their respective original implementations \cite{satorras2021n, huang2022equivariant}. Supervision also varies across models. GNS is trained  to predict  roller accelerations, while EGNN and GMN are supervised on roller positions and velocities. Equi-Euler GraphNet, in contrast,  is only supervised on accelerations of the inner and outer rings. To ensure architectural consistency, all  models use  two-layer MLPs for their node and edge functions.



\section{Results}
\label{sec:results}
\paragraph{Overview}
In this section, we evaluate the performance of the proposed Equi-Euler GraphNet framework on the case study presented in Section~\ref{sec:case_study}, with a focus on two  core  tasks: predicting pairwise contact forces and generating trajectory rollouts for each bearing component in a cylindrical roller bearing.
\paragraph{Model Description}
The Equi-Euler GraphNet takes as input the graph-based representation of the bearing at time \(t\), as described in Section~\ref{sec:graph_rep}, along with the external load and the rotational speed of the inner ring. It evolves the system state over a total time step of  \(5 \times 6.667 \times 10^{-5} \text{ s}\), corresponding  five message-passing iterations. Each iteration represents a sub-time step of \(6.667 \times 10^{-5} \text{ s}\), during which the stacked dynamics and kinematics layers of the Equi-Euler GraphNet are applied (See Section~\ref{sec:model}). The edge and node MLPs consist of two hidden layers with 128 embeddings.

The model predicts the accelerations, updated positions, and velocities of the inner ring and outer ring, the updated positions of the rolling elements, and the pairwise forces between all connected components in the graph. The predicted accelerations and pairwise forces at both \(t\) and \(t+5 \times 6.667 \times 10^{-5} \text{ s}\) are compared against  the corresponding ground-truth values during training.

During rollout, the predicted positions and velocities of the inner ring, outer ring, and rolling elements are used to update the graph state for the next time step, enabling continuous trajectory generation.
\paragraph{Overview of Experiments}
We evaluate the model’s performance across three levels of generalization. In Section~\ref{sec:interp}, we assess interpolation performance on a 13-roller bearing under a 13\,kN load at 600\,RPM. While both the load and roller count are within the training distribution, the rotational speed was not seen during training, making this an interpolation case. The model accurately predicts roller forces over a 6000-step rollout, demonstrating strong physical consistency and stable error accumulation.

In Section~\ref{sec:gen_op_cond}, we test the model’s ability to extrapolate to new operating conditions using a 13-roller bearing at 15\,kN and 1500\,RPM—extending beyond the training rotational speed range. Section~\ref{sec:gen_all} then presents the most challenging extrapolation scenario: a 15-roller bearing operating at 19\,kN and 3000\,RPM—completely outside the training distribution in terms of system configuration and dynamics. Together, these three cases offer a comprehensive evaluation of the model’s ability to interpolate, extrapolate to new operating conditions, and generalize to previously unseen system configurations. Finally, in Section~\ref{sec:res_comp}, we compare the proposed approach with state-of-the-art GNN baselines to establish a quantitative performance benchmark.

\subsection{Performance on Interpolation test case}
\label{sec:interp}
\begin{figure}[h!]  
\centering \includegraphics[width=1.0\textwidth]{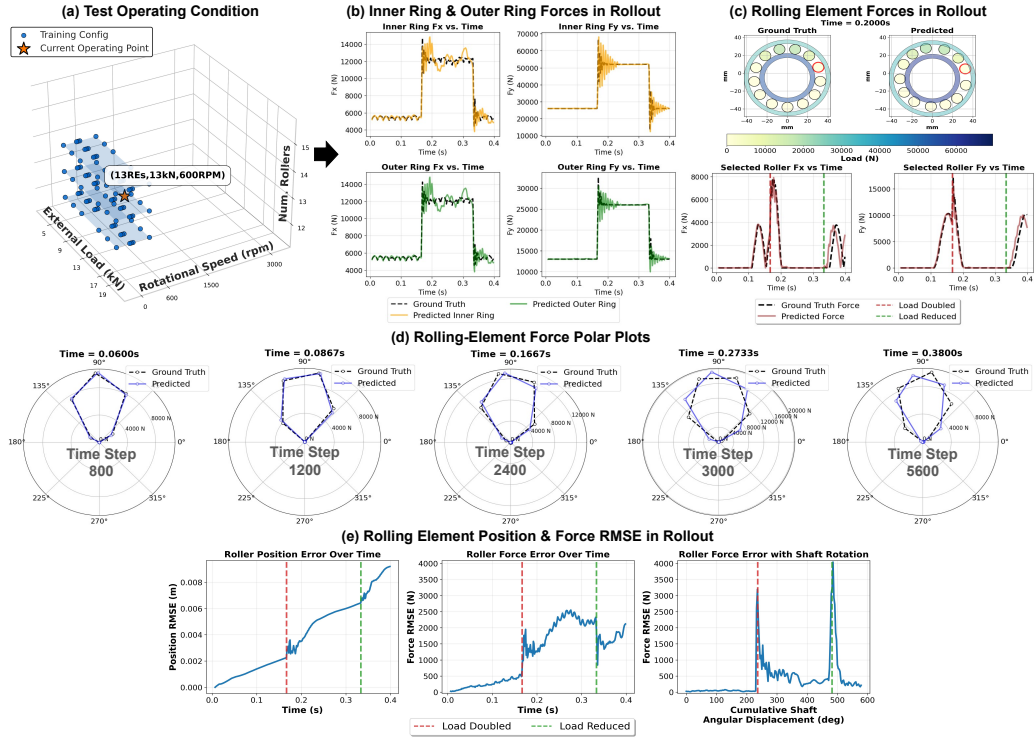}  
\caption{\textbf{Equi-Euler GraphNet Performance on an Interpolation Test Case} (13 rolling elements, 13 kN load, 600 RPM, 5000 time-step rollout):
\textbf{(a)} Location of the test condition relative to the training set.
\textbf{(b)} Predicted vs.\ ground-truth inner and outer ring forces (x and y components) over 5000 time steps, capturing load-doubling and -halving transients.
\textbf{(c)} Rolling-element force evolution for a tracked roller (red outline), with two-dimensional views of all bearing components (ground truth and prediction).
\textbf{(d)} Polar plots (predicted vs.\ ground truth) of roller forces at representative times, illustrating accurate load-distribution predictions, with minor discrepancies arising later in the rollout due to error accumulation.
\textbf{(e)} RMSE of predicted roller position and force over time, and predicted force with cumulative shaft rotation, showing stable error accumulation throughout the extended prediction horizon.}
\label{fig:res_interp}  
\end{figure}

In this experiment, we evaluate Equi-Euler GraphNet’s capability to perform long-range trajectory rollouts and predict pairwise interaction forces under operating conditions drawn from within the training distribution. Specifically, we assess its interpolation performance on a cylindrical roller bearing  with 13 rolling elements, an applied radial load of 13 kN on the outer ring, and an inner ring rotational speed of 600 RPM. Although 600 RPM  was not explicitly used during training, it lies within  the trained rotational speed range. Figure~\ref{fig:res_interp}(a) illustrates this interpolation scenario in relation to the training configurations.

Given the initial graph state, external load, and rotational speed, Equi-Euler GraphNet predicts pairwise contact forces and updates the positions and velocities of the inner ring, outer ring, and rolling elements at each time step. The predicted contact  forces are aggregated to compute the resultant forces acting on  each of these components. Figure~\ref{fig:res_interp}(b) compares the predicted and ground-truth inner and outer ring forces (both $F_x$ and $F_y$ components) over time. Equi-Euler GraphNet accurately captures the forces throughout the first  2500 steps, prior to any load variation. 

At step 2500, when the external load is doubled, the model successfully predicts the resulting transient response. While  the predicted oscillations are qualitatively correct, they exhibit slightly lower damping than the ground truth. Discrepancies are more noticeable in the \(F_x\) component, whereas  the higher-magnitude \(F_y\) forces are more accurately predicted. When the load is subsequently  halved, the model continues   to track the system's response effectively, although the damping behavior remains  slightly under-predicted relative to the ground truth.

Despite minor  discrepancies in the predicted inner and outer ring forces, Figure~\ref{fig:res_interp}(c) shows that Equi-Euler GraphNet accurately captures  the roller loads throughout the extended rollout. The top panel  of the figure depicts  the bearing state at an intermediate time step, highlighting a tracked rolling element (outlined in red) in both the ground truth and predicted configurations. The bottom panel  of the figure shows the load experienced by this tracked roller over time as it traverses  the bearing circumference. The model accurately captures the characteristic load peaks as the roller enters the loaded zone, including a sharp increase  in the \(F_y\) component  following the step change in external load.

To further evaluate  the load distribution among the rolling elements, Figure~\ref{fig:res_interp}(d) presents polar plots of the predicted roller forces at selected rollout time steps. In these plots, the radial distance represents the magnitude of the force acting on each roller, providing a clear visualization of  the loaded zone  as the bearing rotates. The model accurately identifies the location and extent of this zone throughout the rollout. While the overall force patterns remain consistent with the ground truth, minor discrepancies in the distribution emerge  at later time steps, likely due to the gradual accumulation of prediction errors over the extended rollout horizon.

This error accumulation is quantified in Figure~\ref{fig:res_interp}(e). The first two plots show a stable and approximately linear increase in  root mean square error (RMSE) for both position and force over time, indicating that the model exhibits predictable and bounded error growth during long rollouts. As further illustrated in Figure~\ref{fig:res_interp}(c), which displays  the predicted roller loads  throughout the  rollout,  force prediction errors are primarily attributed  to slight positional misalignments of the rollers rather than inaccuracies in the force estimation itself. Although the load  peaks are accurately captured, they appear phase-shifted due to the predicted roller lagging slightly behind the ground truth. To investigate this effect further, the third plot in Figure~\ref{fig:res_interp}(e) presents  the roller force RMSE  a function of cumulative shaft angular displacement, effectively  the roller's cumulative azimuthal position along the bearing circumference as the shaft rotates. This view confirms that the prediction errors remain stable  when aligned with spatial displacement, except during transient load events. This also explains the minor  deviations observed in the polar plots:  although force RMSE increases over time, the spatial  distribution of roller loads remains well aligned  with the ground truth, resulting in consistently accurate identification of the loaded zone.

Overall, this experiment confirms that Equi-Euler GraphNet can reliably perform  extended trajectory rollouts and accurately predict component-wise forces within  range of trained operating conditions, while exhibiting stable and bounded error accumulation over  long  prediction horizons.

\subsection{Extrapolation to an Unseen Operating Condition}
\label{sec:gen_op_cond}
\begin{figure}[h!]  
\centering \includegraphics[width=1.0\textwidth]{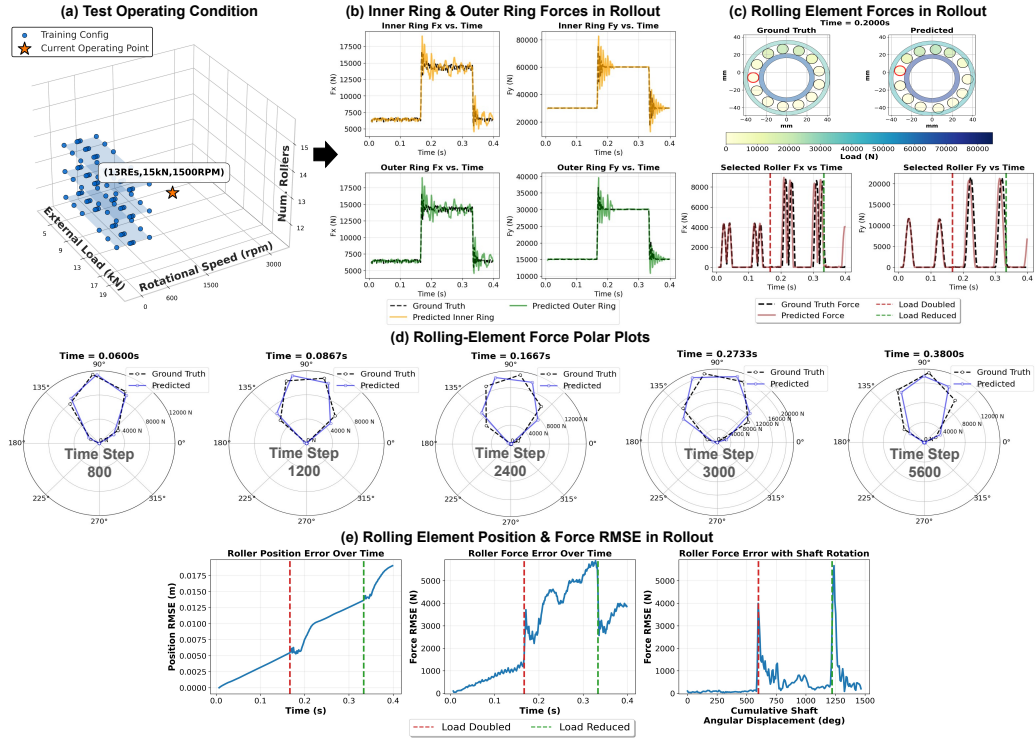}  
\caption{\textbf{Equi-Euler GraphNet Performance on a Generalization Case} (13 rolling elements, 15 kN load, 1500 RPM, 5000 time-step rollout):
\textbf{(a)} Location of the test condition relative to the training set.
\textbf{(b)} Predicted vs.\ ground-truth inner and outer ring forces, capturing steady-state and transient responses at higher-than-trained RPM.
\textbf{(c)} Rolling-element force evolution for a tracked roller, with a 2D snapshot comparing ground truth and prediction.
\textbf{(d)} Polar plots of roller forces at representative times, illustrating accurate load distribution despite minor discrepancies.
\textbf{(e)} RMSE of roller position and force over time and with cumulative shaft rotation, showing stable error growth under unseen operating conditions.}
\label{fig:res_gen}  
\end{figure}

In this experiment, we evaluate  the extrapolation capability of Equi-Euler GraphNet under operating conditions  that extend beyond the training distribution. Specifically, we test the model on a cylindrical roller bearing configuration with 13 rolling elements, subjected to a 15 kN radial load(within the training range) and a rotational speed of 1500 RPM (outside the training RPM range which had a maximum of 750 RPM). Figure~\ref{fig:res_gen}(a) situates  this scenario  in relation  to the training configurations.

Figure~\ref{fig:res_gen}(b) compares the predicted and ground-truth forces on the inner and outer rings. Despite the higher rotational speed and increased dynamic complexity, Equi-Euler GraphNet accurately captures both steady-state forces and transient responses triggered by abrupt load changes  at steps 2500 and 5000. As in the interpolation case, the model displays slightly underdamped behavior, more noticeable  in the \(F_x\) component, while accurately reproducing the dominant \(F_y\) component, that primarily drives the bearing dynamics.

Figure~\ref{fig:res_gen}(c)  presents a snapshot of the predicted and  ground-truth bearing states at an intermediate rollout step, highlighting a tracked roller. The corresponding load on this roller is plotted over time in the lower panel, showing accurate prediction of contact force peaks as the roller enters the loaded zone, including the amplified response after the load doubles at step 2500.

The polar plots in Figure~\ref{fig:res_gen}(d) further confirm that the model maintains accurate spatial load distribution across  multiple time steps.  While minor  deviations appear due to gradual error accumulation during rollout, the overall structure of the loaded zone remains consistent with the ground truth. Figure~\ref{fig:res_gen}(e) quantifies the error accumulation, showing stable and bounded increases  in both position and force RMSE, even under rotational speeds that lie  well beyond  the training distribution. Additionally, the roller force RMSE plotted against cumulative  shaft angular displacement reveals  consistent prediction quality, with noticeable error spikes occurring only during transient load changes. Overall, these results demonstrate that Equi-Euler GraphNet extrapolates well beyond its training regime, accurately predicting  both system trajectories and interaction forces  under operating conditions outside the training distribution.

\subsection{Extrapolation to an Unseen Bearing Configuration, Shaft Rotational Speed and Load}
\label{sec:gen_all}
\begin{figure}[h!]  
\centering \includegraphics[width=1.0\textwidth]{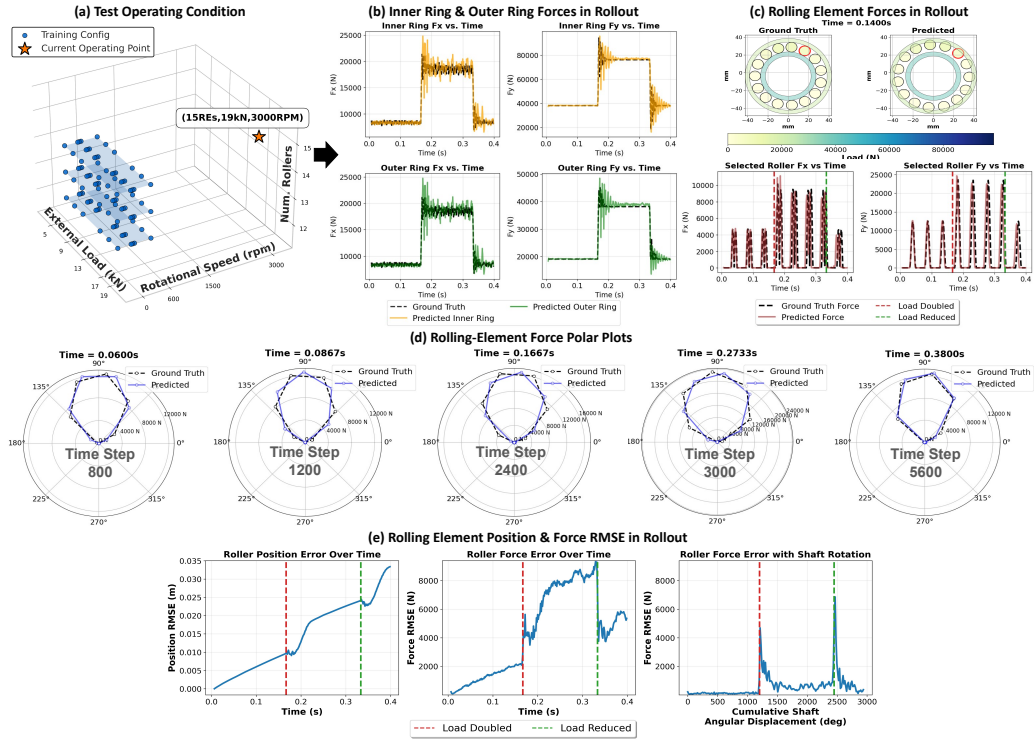}  
\caption{\textbf{Equi-Euler GraphNet Performance on a Challenging Extrapolation Test} (15 rolling elements, 19 kN load, 3000 RPM, 5000 time-step rollout):
\textbf{(a)} Location of the test condition relative to the training set.
\textbf{(b)} Predicted vs.\ ground-truth inner and outer ring forces, showing overall accurate trends but higher-frequency oscillations in the smaller \(F_x\) component.
\textbf{(c)} Rolling-element force evolution for a tracked roller, with a 2D snapshot comparing ground truth and prediction, demonstrating accurate load prediction under untrained operating conditions and bearing configuration.
\textbf{(d)} Polar plots of roller forces at representative times, confirming accurate load distribution despite the new bearing configuration, higher operating speed, and load.
\textbf{(e)} RMSE of roller position and force over time, and force with cumulative shaft rotation, illustrating stable error growth under these significantly beyond-training conditions.}
\label{fig:res_ext}  
\end{figure}
In the most challenging scenario, we evaluate the extrapolation capabilities of Equi-Euler GraphNet  by testing it  on a bearing configuration that lies  well outside the training distribution. This scenario involves a  bearing with  15 rollers (maximum of 13 rollers during training), subjected to an unseen external load of 19 kN (training maximum: 17 kN) and a rotational speed of 3000 RPM (training maximum: 750 RPM)—conditions that significantly exceed those encountered during training  (Figure~\ref{fig:res_ext}(a)).

Figure~\ref{fig:res_ext}(b) compares the predicted and ground-truth forces on the inner and outer rings. While the model accurately captures the overall force trends and transient responses, the predictions exhibit high-frequency oscillations and slightly underdamped behavior compared to the ground truth, reflecting the increased complexity posed by this extrapolation scenario. These discrepancies are more pronounced in the lower-magnitude \(F_x\) component, whereas the dominant  \(F_y\) component, which primarily governs  the system dynamics, is  predicted with high accuracy.

The accurately predicted roller loads shown  in Figure~\ref{fig:res_ext}(c) indicate  that the discrepancies observed  in the \(F_x\) components of the inner and outer rings have minimal influence  on the accuracy of roller load predictions. The  top panel of the figure shows an intermediate snapshot of  the rollout, highlighting a tracked rolling element  (outlined in red)  in both the predicted and ground-truth configurations. The lower panel  plots the contact forces experienced by this 
 roller  over time as  it travels along the bearing circumference. Despite the increased complexity of the extrapolation scenario, the model successfully captures the peak loads as the roller enters the loaded zone,  as well as the amplified  peaks following   the increase in  external load -- demonstrating robust predictive performance at the component level.

Polar force distribution plots at multiple rollout steps (Figure~\ref{fig:res_ext}(d)) confirm that Equi-Euler GraphNet accurately predicts the spatial  distribution of roller loads throughout the extended rollout. The first two plots in Figure~\ref{fig:res_ext}(e) show a stable, yet gradually increasing, trend in position and force RMSE over time, indicating predictable and bounded error accumulation  over  long horizons. The third plot shows the roller force RMSE as a function of  cumulative shaft angular displacement, revealing  consistent  prediction quality relative to each  roller’s azimuthal position along the bearing circumference. These results highlight the model’s ability to maintain spatially coherent and temporally stable force predictions even under highly challenging and previously unseen operating conditions, demonstrating strong extrapolation performance.

\subsection{Comparison Results with Benchmarked GNN Architectures}
\label{sec:res_comp}
\begin{figure}[h!]  
In this section, we compare GNS, EGNN, GMN, and Equi-Euler GraphNet on the interpolation scenario described in Section~\ref{sec:interp}. Since these baseline models show limited performance in this case, we did not extend the comparison to more challenging extrapolation tasks.
To ensure a fair comparison, each baseline model was adapted to predict pairwise forces as detailed in Section~\ref{sec:comp_soa}.

\centering \includegraphics[width=1.0\textwidth]{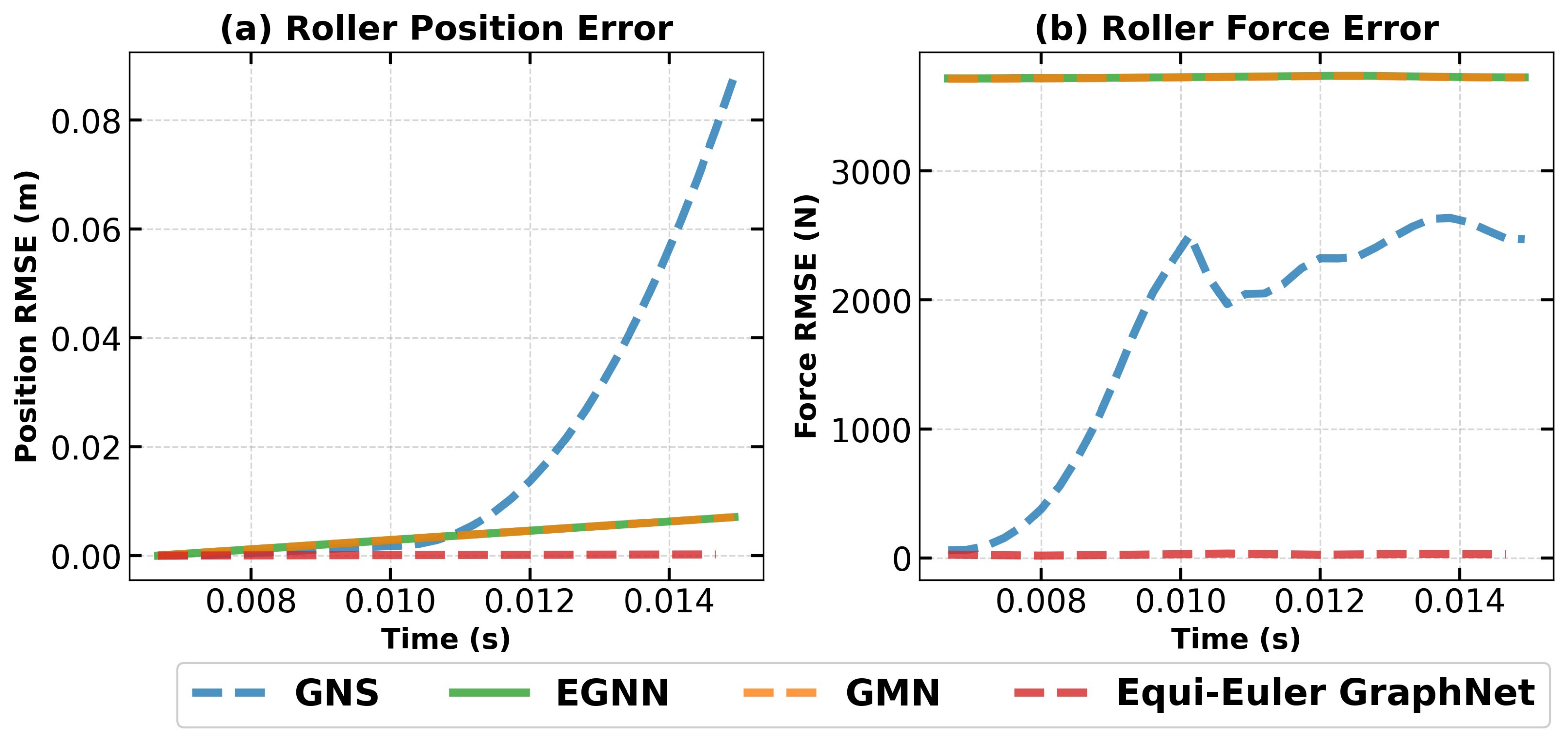}  
\caption{\textbf{Comparison of Trajectory and Force Prediction Accuracy:}:
\textbf{(a)--(b)} Roller position (a) and force (b) root mean square errors over time in an interpolation test (13 rollers, 13 kN load, 600 RPM). Compared to GNS, EGNN, and GMN, \textit{Equi-Euler GraphNet} consistently achieves lower errors in both trajectory and force prediction.}
\label{fig:res_model_comparison}  
\end{figure}

Figure~\ref{fig:res_model_comparison} compares the four models on the interpolation test case involving a 13-roller bearing, under 13\(\,\mathrm{kN}\) load, and 600\(\,\mathrm{RPM}\). Each model is evaluated over a rollout horizon of 225 time steps (0.015 s),  beyond which GNS exhibits prohibitively large errors. Despite having  access to a richer history of past velocities and direct supervision on roller accelerations, GNS quickly accumulates errors in both roller positions and predicted forces. EGNN and GMN produce more stable trajectory rollouts but fail to capture  accurate pairwise interaction forces. In contrast, Equi-Euler GraphNet consistently yields lower errors for both positional and force predictions, demonstrating its capability to jointly learn accurate dynamic trajectories and pairwise interactions. Consequently, Equi-Euler GraphNet outperforms these established approaches for the dual-task needs of trajectory evolution and interaction-force prediction.

\section{Conclusion}
\label{sec:discussion}
In this paper, we propose  Equi-Euler GraphNet, a novel physics-informed graph neural network specifically designed  for virtual sensing and dynamic modeling of cylindrical roller bearings. Our approach  directly tackles  the key  challenge of accurately predicting internal rolling element loads and system trajectories across a wide range of operating conditions--including out-of-distribution bearing configurations and operating conditions.

At the core of  Equi-Euler GraphNet is a unique equivariant message-passing scheme that ensures  predicted internal forces remain consistent under Euclidean transformations.  In parallel, the dynamics-aware  temporal  message passing ensures that the states of components evolve in a physically consistent manner. Our bearing-specific architecture further introduces a tailored structural bias by decoupling the free-body  dynamics of the inner and outer rings from the constrained kinematics of the rolling elements.  This design significantly improves the accuracy of both trajectory predictions and internal force  estimations, while promoting generalization  and stable error accumulation during  extended rollouts.

Through comprehensive  evaluations on high-fidelity simulated data generated using  a multiphysics solver, Equi-Euler GraphNet consistently outperformed state-of-the-art GNN-based baselines, including GNS \cite{sanchez2020learning}, EGNN \cite{satorras2021n}, and GMN \cite{huang2022equivariant}, in both long-horizon trajectory prediction and internal force estimation. Notably, the model demonstrated strong extrapolation capabilities and broad applicability across varying system configurations.  In addition to its accuracy, Equi-Euler GraphNet achieves a substantial computational speedup -- producing extended trajectory rollouts in just 30 seconds, compared to over two hours required by  the multiphysics simulator  under equivalent conditions. These advantages, coupled with its remarkable extrapolation capabilities, position Equi-Euler GraphNet as a compelling reduced-order modeling tool for tasks such as design optimization and digital twin-based condition monitoring through real-time performance evaluation.


Looking ahead, the next key step is to validate Equi-Euler GraphNet on experimental  data from real-world bearing systems. Future work will involve benchmarking  the  model against physical measurements, such as those obtained via embedded load cells in rolling elements \cite{zhao2025graph}, to assess its practical deployment potential. This validation will be critical for transitioning Equi-Euler GraphNet from simulation to real-time  applications, paving the way for integration into industrial predictive maintenance pipelines  for early fault detection and proactive maintenance planning.

\textbf{Acknowledgment of AI Assistance in Manuscript Preparation:}
During the preparation of this work, the authors used ChatGPT to assist with refining and correcting the text. After using this tool, the authors carefully reviewed and edited the content as needed and take full responsibility for the content of this publication.

\bibliographystyle{elsarticle-num}
\bibliography{cas-refs}

\begin{thebibliography}{10}
\expandafter\ifx\csname url\endcsname\relax
  \def\url#1{\texttt{#1}}\fi
\expandafter\ifx\csname urlprefix\endcsname\relax\def\urlprefix{URL }\fi
\expandafter\ifx\csname href\endcsname\relax
  \def\href#1#2{#2} \def\path#1{#1}\fi

\bibitem{cubillo2016review}
A.~Cubillo, S.~Perinpanayagam, M.~Esperon-Miguez, A review of physics-based models in prognostics: Application to gears and bearings of rotating machinery, Advances in Mechanical Engineering 8~(8) (2016) 1687814016664660.

\bibitem{jardine2006review}
A.~K. Jardine, D.~Lin, D.~Banjevic, A review on machinery diagnostics and prognostics implementing condition-based maintenance, Mechanical systems and signal processing 20~(7) (2006) 1483--1510.

\bibitem{li2012fatigue}
S.~Li, A.~Kahraman, M.~Klein, A fatigue model for spur gear contacts operating under mixed elastohydrodynamic lubrication conditions, Journal of mechanical design (1990) 134~(4) (2012).

\bibitem{pandya2013crack}
Y.~Pandya, A.~Parey, Crack behavior in a high contact ratio spur gear tooth and its effect on mesh stiffness, Engineering Failure Analysis 34 (2013) 69--78.

\bibitem{li2017tribo}
S.~Li, A.~Anisetti, A tribo-dynamic contact fatigue model for spur gear pairs, International Journal of Fatigue 98 (2017) 81--91.

\bibitem{qiu2002damage}
J.~Qiu, B.~B. Seth, S.~Y. Liang, C.~Zhang, Damage mechanics approach for bearing lifetime prognostics, Mechanical systems and signal processing 16~(5) (2002) 817--829.

\bibitem{fajdiga2009fatigue}
G.~Fajdiga, M.~Sraml, Fatigue crack initiation and propagation under cyclic contact loading, Engineering fracture mechanics 76~(9) (2009) 1320--1335.

\bibitem{rycerz2017propagation}
P.~Rycerz, A.~Olver, A.~Kadiric, Propagation of surface initiated rolling contact fatigue cracks in bearing steel, International Journal of Fatigue 97 (2017) 29--38.

\bibitem{de2016review}
H.~D.~M. de~Azevedo, A.~M. Ara{\'u}jo, N.~Bouchonneau, A review of wind turbine bearing condition monitoring: State of the art and challenges, Renewable and Sustainable Energy Reviews 56 (2016) 368--379.

\bibitem{morales2019}
G.~E. Morales-Espejel, P.~Engelen, G.~van Nijen, Propagation of large spalls in rolling bearings, Tribology Online 14~(5) (2019) 254--266.

\bibitem{pc1961rational}
P.~PC, A rational analytic theory of fatigue, Trends Engin 13 (1961) 9--14.

\bibitem{meng1995wear}
H.~Meng, K.~Ludema, Wear models and predictive equations: their form and content, Wear 181 (1995) 443--457.

\bibitem{Lundberg1939}
G.~Lundberg, {E}lastische {B}erührung zweier {H}albräume, Forschung auf dem Gebiete des Ingenieurwesens 10~(5) (1939) 201--211.
\newblock \href {https://doi.org/10.1007/bf02584950} {\path{doi:10.1007/bf02584950}}.

\bibitem{palmgren1959}
A.~Palmgren, Ball and roller bearing engineering, Philadelphia: SKF Industries Inc (1959).

\bibitem{Gupta1984}
P.~K. Gupta (Ed.), Advanced Dynamics of Rolling Elements, Springer New York, 1984.

\bibitem{cao2008comprehensive}
M.~Cao, J.~Xiao, A comprehensive dynamic model of double-row spherical roller bearing—model development and case studies on surface defects, preloads, and radial clearance, Mechanical systems and signal processing 22~(2) (2008) 467--489.

\bibitem{cao2018mechanical}
H.~Cao, L.~Niu, S.~Xi, X.~Chen, Mechanical model development of rolling bearing-rotor systems: A review, Mechanical Systems and Signal Processing 102 (2018) 37--58.

\bibitem{li2013tribo}
S.~Li, A.~Kahraman, A tribo-dynamic model of a spur gear pair, Journal of Sound and Vibration 332~(20) (2013) 4963--4978.

\bibitem{ku1998finite}
D.-M. Ku, Finite element analysis of whirl speeds for rotor-bearing systems with internal damping, Mechanical Systems and Signal Processing 12~(5) (1998) 599--610.

\bibitem{wang2005finite}
J.~Wang, I.~Howard, Finite element analysis of high contact ratio spur gears in mesh, J. Trib. 127~(3) (2005) 469--483.

\bibitem{kacprzynski2004predicting}
G.~Kacprzynski, A.~Sarlashkar, M.~Roemer, A.~Hess, B.~Hardman, Predicting remaining life by fusing the physics of failure modeling with diagnostics, JOm 56 (2004) 29--35.

\bibitem{liu2012data}
J.~Liu, W.~Wang, F.~Ma, Y.~Yang, C.~Yang, A data-model-fusion prognostic framework for dynamic system state forecasting, Engineering Applications of Artificial Intelligence 25~(4) (2012) 814--823.

\bibitem{peng2022digital}
F.~Peng, L.~Zheng, Y.~Peng, C.~Fang, X.~Meng, Digital twin for rolling bearings: a review of current simulation and phm techniques, Measurement 201 (2022) 111728.

\bibitem{choi2021data}
H.-S. Choi, J.~An, S.~Han, J.-G. Kim, J.-Y. Jung, J.~Choi, G.~Orzechowski, A.~Mikkola, J.~H. Choi, Data-driven simulation for general-purpose multibody dynamics using deep neural networks, Multibody System Dynamics 51 (2021) 419--454.

\bibitem{zhu2021data}
Y.-P. Zhu, J.~Yuan, Z.-Q. Lang, C.~W. Schwingshackl, L.~Salles, V.~Kadirkamanathan, The data-driven surrogate model-based dynamic design of aeroengine fan systems, Journal of Engineering for Gas Turbines and Power 143~(10) (2021) 101006.

\bibitem{dimitrov2018wind}
N.~Dimitrov, M.~C. Kelly, A.~Vignaroli, J.~Berg, From wind to loads: wind turbine site-specific load estimation with surrogate models trained on high-fidelity load databases, Wind Energy Science 3~(2) (2018) 767--790.

\bibitem{an2015practical}
D.~An, N.~H. Kim, J.-H. Choi, Practical options for selecting data-driven or physics-based prognostics algorithms with reviews, Reliability Engineering \& System Safety 133 (2015) 223--236.

\bibitem{connor1994recurrent}
J.~T. Connor, R.~D. Martin, L.~E. Atlas, Recurrent neural networks and robust time series prediction, IEEE transactions on neural networks 5~(2) (1994) 240--254.

\bibitem{hochreiter1997long}
S.~Hochreiter, J.~Schmidhuber, Long short-term memory, Neural computation 9~(8) (1997) 1735--1780.

\bibitem{lecun1995convolutional}
Y.~LeCun, Y.~Bengio, et~al., Convolutional networks for images, speech, and time series, The handbook of brain theory and neural networks 3361~(10) (1995) 1995.

\bibitem{karniadakis2021physics}
G.~E. Karniadakis, I.~G. Kevrekidis, L.~Lu, P.~Perdikaris, S.~Wang, L.~Yang, Physics-informed machine learning, Nature Reviews Physics 3~(6) (2021) 422--440.

\bibitem{raissi2019physics}
M.~Raissi, P.~Perdikaris, G.~E. Karniadakis, Physics-informed neural networks: A deep learning framework for solving forward and inverse problems involving nonlinear partial differential equations, Journal of Computational physics 378 (2019) 686--707.

\bibitem{zhao2024comprehensive}
C.~Zhao, F.~Zhang, W.~Lou, X.~Wang, J.~Yang, A comprehensive review of advances in physics-informed neural networks and their applications in complex fluid dynamics, Physics of Fluids 36~(10) (2024).

\bibitem{haghighat2021physics}
E.~Haghighat, M.~Raissi, A.~Moure, H.~Gomez, R.~Juanes, A physics-informed deep learning framework for inversion and surrogate modeling in solid mechanics, Computer Methods in Applied Mechanics and Engineering 379 (2021) 113741.

\bibitem{new2023tunable}
A.~New, B.~Eng, A.~C. Timm, A.~S. Gearhart, Tunable complexity benchmarks for evaluating physics-informed neural networks on coupled ordinary differential equations, in: 2023 57th Annual Conference on Information Sciences and Systems (CISS), IEEE, 2023, pp. 1--8.

\bibitem{li2024physics}
Z.~Li, H.~Zheng, N.~Kovachki, D.~Jin, H.~Chen, B.~Liu, K.~Azizzadenesheli, A.~Anandkumar, Physics-informed neural operator for learning partial differential equations, ACM/JMS Journal of Data Science 1~(3) (2024) 1--27.

\bibitem{krishnapriyan2021characterizing}
A.~Krishnapriyan, A.~Gholami, S.~Zhe, R.~Kirby, M.~W. Mahoney, Characterizing possible failure modes in physics-informed neural networks, Advances in neural information processing systems 34 (2021) 26548--26560.

\bibitem{battaglia2018relational}
P.~W. Battaglia, J.~B. Hamrick, V.~Bapst, A.~Sanchez-Gonzalez, V.~Zambaldi, M.~Malinowski, A.~Tacchetti, D.~Raposo, A.~Santoro, R.~Faulkner, et~al., Relational inductive biases, deep learning, and graph networks, arXiv preprint arXiv:1806.01261 (2018).

\bibitem{battaglia2016interaction}
P.~Battaglia, R.~Pascanu, M.~Lai, D.~Jimenez~Rezende, et~al., Interaction networks for learning about objects, relations and physics, Advances in neural information processing systems 29 (2016).

\bibitem{mrowca2018flexible}
D.~Mrowca, C.~Zhuang, E.~Wang, N.~Haber, L.~F. Fei-Fei, J.~Tenenbaum, D.~L. Yamins, Flexible neural representation for physics prediction, Advances in neural information processing systems 31 (2018).

\bibitem{sanchez2020learning}
A.~Sanchez-Gonzalez, J.~Godwin, T.~Pfaff, R.~Ying, J.~Leskovec, P.~Battaglia, Learning to simulate complex physics with graph networks, in: International conference on machine learning, PMLR, 2020, pp. 8459--8468.

\bibitem{choi2024graph}
Y.~Choi, K.~Kumar, Graph neural network-based surrogate model for granular flows, Computers and Geotechnics 166 (2024) 106015.

\bibitem{atz2021geometric}
K.~Atz, F.~Grisoni, G.~Schneider, Geometric deep learning on molecular representations, Nature Machine Intelligence 3~(12) (2021) 1023--1032.

\bibitem{thangamuthu2022unravelling}
A.~Thangamuthu, G.~Kumar, S.~Bishnoi, R.~Bhattoo, N.~Krishnan, S.~Ranu, Unravelling the performance of physics-informed graph neural networks for dynamical systems, Advances in Neural Information Processing Systems 35 (2022) 3691--3702.

\bibitem{satorras2021n}
V.~G. Satorras, E.~Hoogeboom, M.~Welling, E (n) equivariant graph neural networks, in: International conference on machine learning, PMLR, 2021, pp. 9323--9332.

\bibitem{huang2022equivariant}
W.~Huang, J.~Han, Y.~Rong, T.~Xu, F.~Sun, J.~Huang, \href{https://openreview.net/forum?id=SHbhHHfePhP}{Equivariant graph mechanics networks with constraints}, in: International Conference on Learning Representations, 2022.
\newline\urlprefix\url{https://openreview.net/forum?id=SHbhHHfePhP}

\bibitem{han2022equivariant}
J.~Han, W.~Huang, T.~Xu, Y.~Rong, Equivariant graph hierarchy-based neural networks, Advances in Neural Information Processing Systems 35 (2022) 9176--9187.

\bibitem{sharma2025dynami}
V.~Sharma, O.~Fink, Dynami-cal graphnet: A physics-informed graph neural network conserving linear and angular momentum for dynamical systems, arXiv preprint arXiv:2501.07373 (2025).

\bibitem{wang2021}
Q.~Wang, C.~Taal, O.~Fink, Integrating expert knowledge with domain adaptation for unsupervised fault diagnosis, IEEE Transactions on Instrumentation and Measurement 71 (2022) 1--12.

\bibitem{han2022learning}
J.~Han, W.~Huang, H.~Ma, J.~Li, J.~Tenenbaum, C.~Gan, Learning physical dynamics with subequivariant graph neural networks, Advances in Neural Information Processing Systems 35 (2022) 26256--26268.

\bibitem{Gupta1979}
P.~K. Gupta, Dynamics of rolling-element bearings - part i: Cylindrical roller bearing analysis, Journal of Lubrication Technology 101~(3) (1979) 293--302.
\newblock \href {https://doi.org/10.1007/978-1-4612-5276-4} {\path{doi:10.1007/978-1-4612-5276-4}}.

\bibitem{zhao2025graph}
M.~Zhao, C.~Taal, S.~Baggerohr, O.~Fink, Graph neural networks for virtual sensing in complex systems: Addressing heterogeneous temporal dynamics, Mechanical Systems and Signal Processing 230 (2025) 112544.

\bibitem{tsuha2020stiffness}
N.~A.~H. Tsuha, K.~L. Cavalca, Stiffness and damping of elastohydrodynamic line contact applied to cylindrical roller bearing dynamic model, Journal of Sound and Vibration 481 (2020) 115444.

\bibitem{sharma2023graph}
V.~Sharma, J.~Ravesloot, C.~Taal, O.~Fink, Graph neural networks for dynamic modeling of roller bearings, in: Annual Conference of the PHM Society, Vol.~15, 2023.

\bibitem{brandstetter2022message}
J.~Brandstetter, D.~E. Worrall, M.~Welling, \href{https://openreview.net/forum?id=vSix3HPYKSU}{Message passing neural {PDE} solvers}, in: International Conference on Learning Representations, 2022.
\newline\urlprefix\url{https://openreview.net/forum?id=vSix3HPYKSU}

\end{thebibliography}

\end{document}